# The Neural Network Pushdown Automaton: Model, Stack and Learning Simulations

G.Z. Sun[1,2], C.L. Giles[2,3], H.H. Chen[1,2] and Y.C. Lee[1,2]

[1]Laboratory For Plasma Research,
[2]Institute for Advanced Computer Studies
University of Maryland, College Park, MD 20742
and
[3]NEC Research Institute
4 Independence Way, Princeton, NJ 08540

sun@sunext.umiacs.umd.edu

## Abstract

In order for neural networks to learn complex languages or grammars, they must have sufficient computational power or resources to recognize or generate such languages. Though many approaches have been discussed, one obvious approach to enhancing the processing power of a recurrent neural network is to couple it with an external stack memory - in effect creating a neural network pushdown automata (NNPDA). This paper discusses in detail this NNPDA - its construction, how it can be trained and how useful symbolic information can be extracted from the trained network.

In order to couple the external stack to the neural network, an optimization method is developed which uses an error function that connects the learning of the state automaton of the neural network to the learning of the operation of the external stack. To minimize the error function using gradient descent learning, an analog stack is designed such that the action and storage of information in the stack are continuous. One interpretation of a continuous stack is the probabilistic storage of and action on data. After training on sample strings of an unknown source grammar, a quantization procedure extracts from the analog stack and neural network a discrete pushdown automata (PDA). Simulations show that in learning deterministic context-free grammars - the balanced parenthesis language, $1^n0^n$, and the deterministic Palindrome - the extracted PDA is *correct* in the sense that it can correctly recognize unseen strings of arbitrary length. In addition, the extracted PDAs can be shown to be identical or equivalent to the PDAs of the source grammars which were used to generate the training strings.

## I. INTRODUCTION

Recurrent neural networks are dynamical network structures which have the capabilities of processing and generating temporal information. To our knowledge the earliest neural network model that processed temporal information was that of McCulloch and Pitts [McCulloch43]. Kleene [Kleene56] extended this work to show the equivalence of finite automata and McCulloch and Pitts' representation of nerve net activity. Minsky [Minsky67 showed that any hard-threshold neural network could represent a finite state automata and developed a method for actually constructing a neural network finite state automata. However, many different neural network models can be defined as recurrent; for example see [Grossberg82] and [Hopfield82]. Our focus is on discrete-time recurrent neural networks that dynamically process temporal information and follows in the tradition of recurrent network models ini-tially defined by [Jordan86] and more recently by [Elman90] and [Pollack91]. In particular this paper develops a neural network pushdown automaton (NNPDA), a *hybrid* system that couples a recurrent network to an external stack memory. More importantly, a NNPDA should be capable of learning and recognizing some class of Context-free grammars. As such, this model is a significant extension of previous work where neural network finite state automata simulated and learned regular grammars. We explore the capabilities of such a model by inferring automata from sample strings - the problem of grammatical inference. It is important to note that our focus is only on that of inference, not of prediction or translation. We will be concerned with problem of inferring an unknown system model based on observing sample strings and not on predicting the next string element in a sequence.

## 1.1 Motivation

To enhance the computational power of a recurrent neural network finite state automaton to that of an *infinite machine* [Minsky67] requires an expansion of resources. One way to achieve this goal is to introduce a potentially infinite number of neurons but a finite set of uniformly distributed local connection weights per neuron. [Sun91] is an example of this approach and shows the Turing equivalence by construction. Another way to construct a neural network *infinite machine* is to allow infinite precision of neuron units but keep a finite size network (finite number of neurons and connection weights) [Siegelmann91, Pollack87]. Doing so is equivalent to constructing a more general nonlinear dynamic system with a set of continuous, recurrent state variables. Such a system in general would have rich dynamical behavior: fixed points, limit cycles, strange attractors and chaos, etc. However, how is such a system trained? In general, without additional knowledge it is almost impossible to train an infinite neural system to learn a desired behavior. In effect, putting constraints and *a priori* knowledge in learning systems has been shown to significantly enhance the practical capabilities of those systems. The model we introduce has this flavor. It enhances the neural network by giving the neural network an infinite memory - a stack - and constrains the learning model by permitting the network to operate on the stack in the standard pre-specified way - *push*, *pop* or *no-operation* (*no-op*). As such, this model can be viewed as: (1) a neural network system with some special constrains on an infinite neural memory, or (2) a hybrid system which couples an external stack memory (conventionally a discrete memory, but here a continuous stack) with a finite size neural network state automaton.

## 1.2 Grammars and Grammatical Inference

Because this paper is concerned with new models neural networks, we give only a brief explanation of grammars and grammatical inference. For more details, please see the enclosed references. Grammatical inference is the problem of inferring an unknown grammar from only grammatical string samples [Angluin83, Fu82, Gold78, Miclet90]. In the Chomsky hierarchy of phrase structured grammars [Harrison78, Hopcroft79, Partee90], the simplest grammars and its associated automata are regular grammars and finite state automata (FSA). Moving up in complexity in the Chomsky hierarchy, the next class is the context-free grammars (CFGs) and their associated recognizer - the pushdown automata (PDA), where a finite state automaton has to control an external stack memory in addition to its own state transition rules. For all classes of grammars, the grammatical inference problem is in the worst case at least NP [Angluin83]. Because of the difficulty of this problem, we feel that training a neural network to learn grammars is a good testbed for exploring the networks computational capabilities.

## 1.3 Outline of Paper

In next section, we review some of the previous work on recurrent neural network finite state automata and work that extends the power of recurrent neural network beyond that of a finite state automata. We show that from the standpoint of representation, it is more computationally efficient to use a "real" external stack instead of the neural network emulator of stack memory [Pollack90]. In Section III we systematically introduce the model of the Neural Network Pushdown Automata (NNPDA), the structure, the dynamics and the optimization (learning) algorithms. This model is substantiated by means of theoretical analysis of many of the related issues regarding its construction. The attempt there is to give a rigorous mathematical description of the NNPDA structure. We then illustrate the model by correctly learning the context-free languages: balanced parentheses and the $\mathbf{1^n\ 0^n}$. A modified version of NNPDA is then introduced to learn the more difficult Palindrome grammar. The conclusion covers enhancements and further directions. In the Appendices, a detailed mathematical derivation of the crucial formula necessary for the training equations of NNPDA is discussed. The key point is that in order to use real-time recurrent learning (RTRL) algorithm [Williams89], we have to assume a recursion relation for all variables, which means that the NNPDA model must be approximated by a finite state automaton. In the Appendices, we discuss this paradox and show one solution to this problem.

# II. RELATED WORK

In this section we review previous work related to the NNPDA. However, the general area of grammatical inference and language processing will not be covered; see for example [Angluin83, Fu82, Miclet90] and more recently the proceedings of the workshop on grammatical inference [Lucas93]. We only focus on neural network related research and, even there, only on work directly related to our model.

## 2.1 Recurrent Neural Network - Connectionist State Machine

Recurrent neural networks have been explored as models for representing and learning formal and natural languages. The basic structure of the recurrent networks, shown in Fig. 1, is that of a neural network finite state automaton (NNFSA) [Allen90, Cleeremans89, Giles92a, Horne92, Liu90, Mozer90, Noda92, Pollack91, Sanfeliu92, Watrous92]. More recently, [Nerrand93] formalizes recurrent networks in a finite-state canonical form. We will not directly discuss neural network finite state machines, i.e. NNFSA which have additional output symbols, see for example [Das91, Chen92]. The computational capabilities of recurrent networks were discussed more recently by [Giles92a, Pollack91, Siegelmann92].

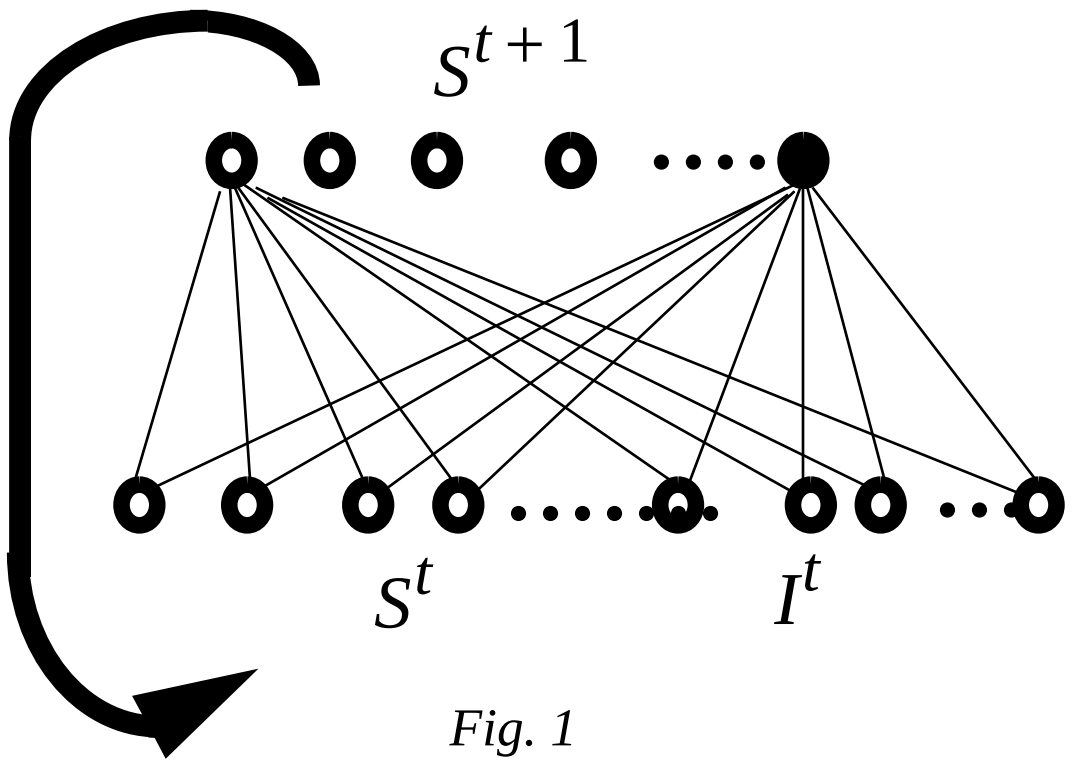

*Fig. 1*

Fig.1 A simple structure of a recurrent neural network, where $I^t$ and $S^t$ represent the current input and state, and $S^{t+1}$ is the next state.

All of the recurrent network models discussed will be higher-order. We and others have found that these models can be extremely useful and more powerful for representing specific computational constructs in neural networks; for a discussion of their use see the following papers [Lee86, Goudreau93, Miller93, Pao89, Perantonis92, Pollack87, Psaltis88, Watrous92]. (It is easy to see that higher order terms are more general than sigma-pi [Rumelhart86a] or pi-sigma [Ghosh92] expressions.) Using second order connection weights, the recurrent dynamics of the state neurons can be given by

$$S_i^{t+1} = g\left(\sum_{j,k} W_{ijk} S_j^t I_k^t + \theta_i\right) , \qquad (1)$$

where $S_i^t$ is the activity of the $i_{th}$ *State* neuron at time step $t$, $I_k^t$ is the $k_{th}$ component of the input symbol at time step t, $g$ is the nonlinear operator, usually the sigmoid function $g(x) = 1 / (1+exp(-x))$ and $\theta_i$ is the bias term for the $i_{th}$ neuron. When a temporal sequence of length $T$: $\{\mathbf{I}^1, \mathbf{I}^2, \mathbf{I}^3, ......, \mathbf{I}^T\}$ is fed into the recurrent net, the input symbol $\mathbf{I}^t$ at each time step together with the current state $\boldsymbol{S}^t$ (initial state is assigned) are the “input” to the network and the “output” would be the next time state $\boldsymbol{S}^{t+1}$. The recurrent network therefore acts like a state automata. At the end of an input string, an end symbol is given to the network and the output in the last state neuron is checked to determine the classification category of the input string. This neural network finite state automaton (NNFSA) can be used to recognize strings that belong to a regular grammar. The work of [Cleeresman89, Giles92a, Giles92b, Liu90, Omlin92, Pollack91, Wa-

trous92, Zeng93] has shown the possibility of using neural networks to perform grammatical inference on regular grammar, i.e. to find a "useful set" of production rules *P* from only a finite set of sample training strings.

One of the limitations of NNFSA is its difficulty in processing higher level languages. A "brute-force" method to enhance the computational power of a NNFSA is to increase the size of the existing neural network structure (or increase the precision of the neuron units in the network) while training on a more complex language, say a context-free grammar [Allen90]. The assumption is that the size of the neural networks has no bound, but the knowledge gained as the network grows gives clues to the representation of the underlying grammar and it associated machine ([Crutchfield91] uses this approach to show that context-free grammars are generated by a nonlinear system on the edge of chaos). But in practice gaining this knowledge is difficult. What usually happens is that the trained NNFSA will only recognize the language up to a certain string length (in effect, a regular grammar). For the NNFSA to generalize correctly on longer unseen strings, the NNFSA needs to be re-trained on those strings. Thus, we argue that this method of knowledge representation is in itself inefficient.

## 2.2 Recurrent Neural Network - Beyond the Finite State Automaton

There has been a great deal of effort to enhance the power of recurrent neural networks by increasing the precision or size of the network or by coupling it with an external, potentially infinite, memory. The work of [Williams89] coupled a recurrent neural network to a memory tape to emulate a Turing machine and to learn the state automaton controller for the balanced-parentheses grammar (a context-free grammar). More specifically, a recurrent network was trained to be the correct finite-state controller of a given Turing machine by supervising the input-output pairs, where the input is the tape reading from a target Turing machine and the output is the desired action of the finite controller. The important distinction between NNPDA model and that of [Williams89] is in the training - particularly, the behavior of their target controller was known *a priori and not learned*. In the most general case of grammatical inference the transition rules of the target machine are not known beforehand; only the classification for each training sequence is known. The NNPDA model we describe allows the NNPDA itself to "figure out" how to construct a neural net controller that knows both the state transition rules and, in addition, how to use and manipulate the tape or stack.

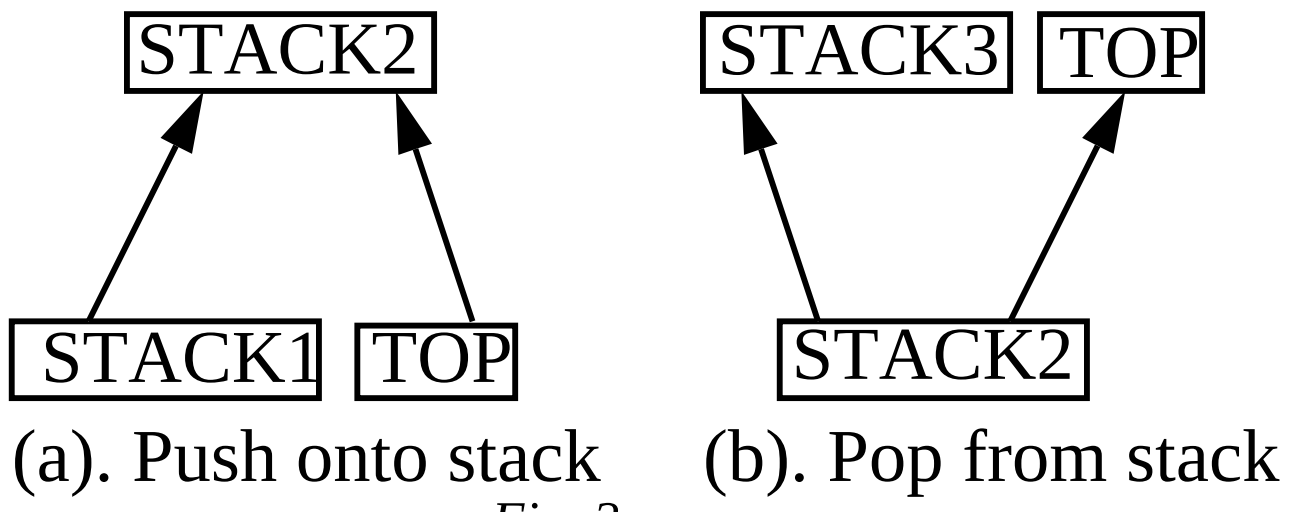

*Fig. 2*

Fig.2 A neural network emulator of a stack proposed by [Pollack90]. (a) Coding process emulates a "push" action onto a stack. (b) Decoding process emulates a "pop" action from a stack.

Closely related work is the RAAM model of [Pollack90], which proposed an "internal" neural network model of stack memory as a plausible model for cognitive processing. Let us consider using this model to build a NNPDA. As shown in Fig. 2, the "push" and "pop" actions onto the stack are emulated by a coder and a decoder separately, where the "STACK1", "STACK2", and "STACK3" are the neuron arrays with the same size and the "TOP" represents the symbol(s) on the top of the stack. The training can be performed by concatenating the network in Fig2(b) with the network in Fig2(a) and using error back-propagation. The desired outcome requires "STACK3" to be identical to "STACK1". This recursive distributed representation of a stack memory may be of particular interest to cognitive models of language processing. However, as a computational model this structure has drawbacks. First, this recursive structure is identical to a NNFSA, where the "STACK's" configurations correspond to internal neural states. In other words, this model transfers the complexity of a stack manipulation to NNFSA state transitions. For a stack with limited length, this model is equivalent to training a FSA with a small number of states. But in general, such a model will be limited since, theoretically, the stack represents a potentially infinite number of states. Even for a limited length stack, this model is inefficient. To illustrate this, consider a stack with length L and number of symbols N. The total number of possible configurations of the stack is

$$N_s = \sum_{l=0}^{L} N^l = \frac{N^{L+1} - 1}{N - 1} \sim N^L \ . \tag{2}$$

If we wish to build a distributed memory of internal states that behaves like a stack, we need to construct (or learn) a NNFSA with $N^L$ internal states. The required memory size of neurons (or weights) will scale as $\sim N^L$ which severely limits the usefulness of the internal neural network stack.

Other closely related work is the connectionist Turing machine models of [Siegalmann92, Pollack87]. They showed that a stack can be simulated in terms of binary representations of a fractional number which are manipulated by neural network generated actions. The focus of this work was initially on "representational" issues and not on a "practical" learning system. Their proposed stacks use a fractional number represented in terms of a sequence of binary symbols "0" and "1". A "pop" action removes the leading bit from the fraction and can be simulated by two consecutive numerical operations: multiplication by two and subtraction of the leading bit. A "push" is represented by adding "0" or "1" to the original stack and dividing the sum by two. This stack model is clearly as efficient as the conventional discrete stack. An additional feature is its simple representation -- a fractional number. However, for learning these stack models have the problem that they are not easy to couple to gradient-based learning algorithms. This is because, although a fractional number is continuous, any small perturbation of the fraction causes a discrete change of the stack content that this fraction is representing.

Finally, an interesting model developed by [Lucas90] proposes an entirely different method for learning context-free grammars with a neural network. [Lucas90] maps directly the production rules of the CFG, both terminals and nonterminals, directly in neural networks and shows some preliminary results for character recognition. ([Frasconi93, Giles93, Sanfeliu92] illustrate similar techniques for mapping regular grammars into recurrent networks.)

The original NNPDA model with an external continuous stack and its learning algorithm were originally proposed in short papers [Giles90, Sun90a, Sun90b]. Recently [Das92] showed benchmark experiments with different order connection weights of NNPDA and pointed out that third order weights were better than first or second orders. [Das93] showed the advantage of using hints in learning CFGs. Recent work of [Mozer93] also shows that the continuous stack can be used to manipulate the "continuous rewrite rules" necessary to parse context-free grammars.

# III. NEURAL NETWORK PUSHDOWN AUTOMATA

In this section, the NNPDA model is thoroughly described. The schematic diagram of the neural network pushdown automata (NNPDA) is shown in Fig. 3. This NNPDA, after being trained, will hopefully be able to represent the underlying grammar of the given training set (we assume that for each of our training sets there is a unique underlying grammar) and be able to correctly classify all unseen input strings generated by an unknown CFG. To use the NNPDA as a classifier, input strings are fed into the NNPDA one character a time, and the "error function" at the end of each string sequence decides the classification. It is important to note that all grammars and automata discussed in this paper are deterministic.

The proposed NNPDA consists of two major components: a recurrent neural network controller and an external continuous stack memory. The structure and working mechanism of these two components will be described in detail in subsections 3.1 and 3.2. A brief introduction of the NNPDA dynamics follows. The neural network controller consists of four types of neurons: input neurons, state neurons, action neurons and stack reading neurons; and the stack is simply a conventional stack with analog symbol "length". At each time step, the recurrent neural network can be considered an input-output mapping. The input to the mapping is: the current internal state $\mathbf{S}^t$, input symbol $\mathbf{I}^t$ and the stack reading $\mathbf{R}^t$. And the output are the next time internal state $\mathbf{S}^{t+1}$ and the stack action $\mathbf{A}^{t+1}$. This action will be performed onto the external stack, which in turn will renew the next time stack reading $\mathbf{R}^{t+1}$. This new stack reading together with new internal state $\mathbf{S}^{t+1}$ and new input symbol $\mathbf{I}^{t+1}$ will serve as a new input for another input-output mapping. At the end of input sequence the content of internal state and stack will determine whether or not the input string is legal.

During the training stage, the weights of the recurrent neural net will be modified to minimize the error function, which is fully discussed in subsections 3.4 and 3.5. In sense the learning can be thought of as unsupervised or reinforcement style learning, because (a) no credit assignment is made before the end of input sequences and (b) the system can extract the classification rules automatically from the input examples.

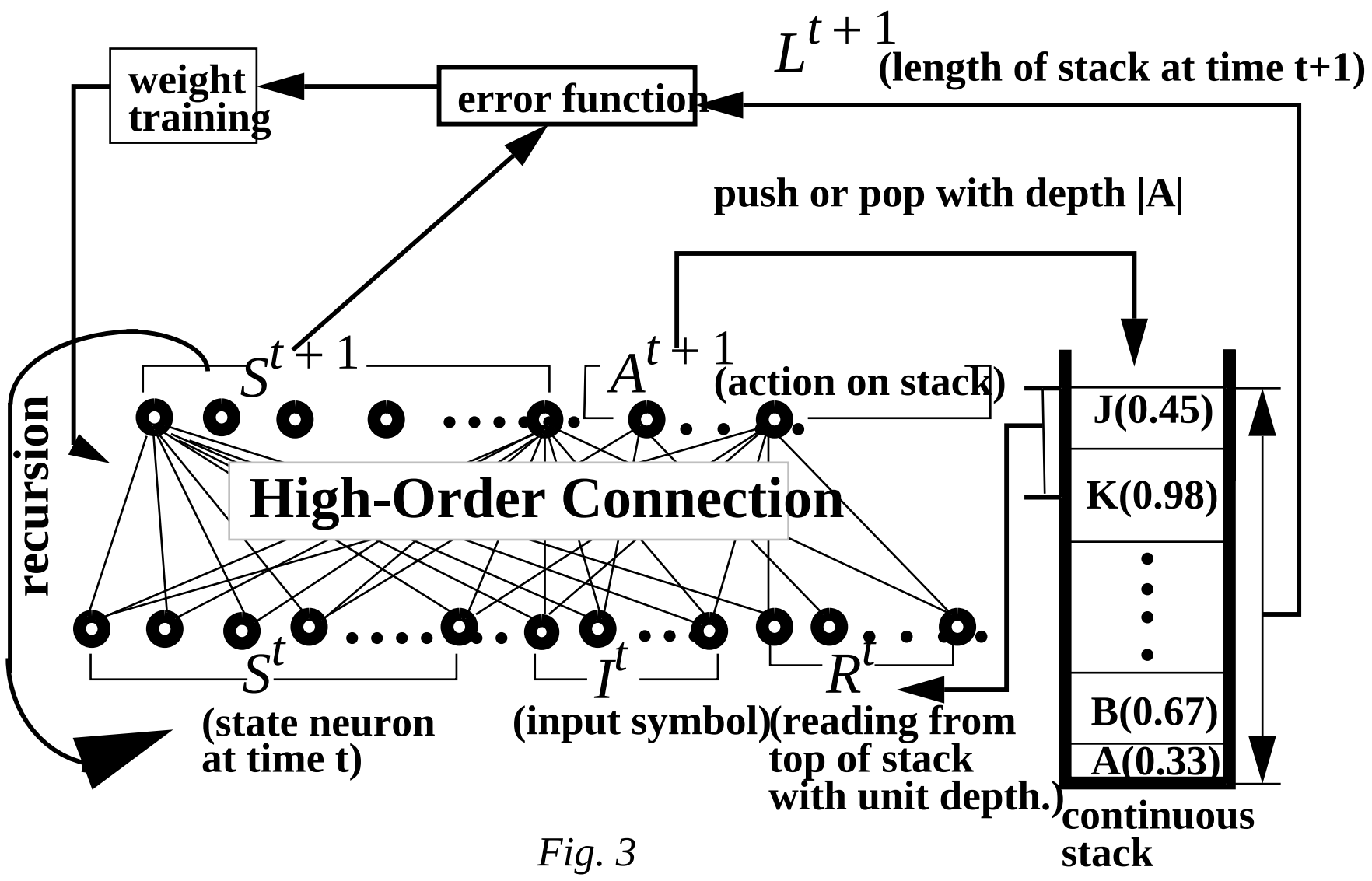

*Fig. 3*

Fig.3 The schematic diagram of the Neural Network Pushdown Automata NNPDA, where a high-order recurrent network is coupled with an external continuous stack. The inputs to the neural net are the current internal states ($S^t$), input symbols ($I^t$) and the stack reading ($R^t$). The outputs from the neural net are the next time internal state ($S^{t+1}$) and the stack action ($A^{t+1}$). This action will be performed on the external stack, which in turn will renew the next stack reading ($R^{t+1}$). The weights of the recurrent neural network controller will be trained by minimizing the error function, which is a function of the final state and the stack length at the end of input string.

## 3.1 Neural Network Controller

The neural network controller is an extended version of the neural network finite state automata (NNFSA) previously described in [Giles92a, Liu90]. It is still a high order recurrent neural network (Fig.3). The difference is that the NNPDA introduces additional input and output neurons (and, of course, the external stack). The “hidden” recurrent neurons $\{S_i, i=1,2,...,N_S\}$ represent the internal states of the system to be learned. The input neurons $\{I_i, i=1,2,...,N_I\}$, are each associated with a particular input symbol (a localist or one-hot encoding scheme). These two groups of neurons are the same as that of NNFSA. The additional “nonrecurrent” input neurons $\{R_i, i=1,2,...,N_R\}$ represent the stack content read from the top of stack memory. The additional “nonrecurrent” output neurons $\{A_i, i=1,2,...,N_A\}$ represent the action values that operate the stack (*pushes*, *pops* or *no-operations*). The state neurons are feedback into themselves after one time step delay (Fig. 3).

The discrete time dynamics of the neural network controller can be written in general form as

$$\begin{aligned} \boldsymbol{S}^{t+1} &= G(\boldsymbol{S}^t, \boldsymbol{R}^t, \boldsymbol{I}^t; \boldsymbol{W}^s) \\ \boldsymbol{A}^{t+1} &= F(\boldsymbol{S}^t, \boldsymbol{R}^t, \boldsymbol{I}^t; \boldsymbol{W}^a) \end{aligned}, \qquad (3)$$

where $\boldsymbol{S}^t$, $\boldsymbol{R}^t$ and $\boldsymbol{I}^t$ are vectors of internal state, stack reading and input symbol at time $t$, and $\boldsymbol{W}^s$ and $\boldsymbol{W}^a$ represent the weight matrices for the state dynamics and action mappings. It is seen from Eq.(3) that for a full description of the dynamic, we need another equation for the stack reading $\boldsymbol{R}^t$. In general, this function could be written as

$$\boldsymbol{R}^t = F(\boldsymbol{A}^1, \boldsymbol{A}^2, \ldots, \boldsymbol{A}^t, \boldsymbol{I}^1, \boldsymbol{I}^2, \ldots, \boldsymbol{I}^t) \, . \qquad (4)$$

Combining Eqs. (3) and (4), for a given set of initial values of $\boldsymbol{S}^0$, $\boldsymbol{R}^0$ and $\boldsymbol{A}^0$, the system “state variables” $\{\boldsymbol{S}^t, \boldsymbol{R}^t, \boldsymbol{A}^t\}$ will evolve in time as an input sequence $\{\boldsymbol{I}^1, \boldsymbol{I}^2, \boldsymbol{I}^3,......,\boldsymbol{I}^T\}$ is fed in. However, this is not a state machine, because Eq.(4) indicates that there does not exist a simple recursive function for the stack reading $\boldsymbol{R}^t$. The value of $\boldsymbol{R}^t$ depends on the entire history of input and actions (or equivalently, $\boldsymbol{R}^t$ depends on weight matrices and input history). This mapping of $\boldsymbol{R}^t$ is highly nonlinear and is determined by the definition of the stack mechanism, which will be later discussed in detail. To be exact, the so called neural network controller is defined only by Eq.(3).

To decide the proper structure of neural network controller, both the neural representations and the target mapping functions need to be known. For discrete pushdown automata, the mappings (or transition rules) are third-order in nature, by which we mean that each transition rule is a unique mapping from a third-order combination: $\{\boldsymbol{S}^t \times \boldsymbol{R}^t \times \boldsymbol{I}^t\}$ to its output, the next time state $\boldsymbol{S}^{t+1}$ and stack action $\boldsymbol{A}^{t+1}$. Assume that unary representations of $\boldsymbol{I}^t$, $\boldsymbol{R}^t$ and $\boldsymbol{S}^t$ are employed. For instance let $\boldsymbol{I}^t$=(1, 0, 0), (0, 1, 0) and (0, 0, 1) represent symbols ***a, b*** and ***c***, and $\boldsymbol{S}^t$ =(1, 0) and (0, 1) the two different states. It is easily seen that any transition rule: $\{S_j^t, R_k^t, I_l^t\} \rightarrow S_i^{t+1}$ or $A_i^{t+1}$ could be coded into two four-dimensional matrices $W^s_{ijkl}$ and $W^a_{ijkl}$, each component being a binary value 0 or 1(for $W^s_{ijkl}$), or ternary value 1, 0, -1(for $W^a_{ijkl}$). For example, the state transition rule $\{\boldsymbol{S(j)}, \boldsymbol{R(k)}, \boldsymbol{I(l)}\} \rightarrow \boldsymbol{S(i)}$ means that if the input symbol is the $l_{th}$ symbol, the stack reading is the $k_{th}$ symbol and the internal state is the $j_{th}$ state, then the next state will be the $i_{th}$ state. And, this rule would be coded as $W^s_{ijkl}$=1 and $W^s_{mjkl}$=0, m≠i. Similarly, $W^a_{ijkl}$= [1, 0, -1] implies a mapped action: [*push*, *no-op*, *pop*] of $A_i^{t+1}$. In this way we show that any deterministic PDA could be implemented by a third order, one layer recurrent neural network with discrete neural activity function. Particularly, if the NNPDA's neural network controller is represented by third-order nets of the form

$$\begin{aligned} S_i^{t+1} &= g\Big(\sum_{j,k,l} W^s_{ijkl}\,(S_j^t R_k^t I_l^t) + \theta_i^s\Big) \\ A_i^{t+1} &= f\Big(\sum_{j,k,l} W^a_{ijkl}\,(S_j^t R_k^t I_l^t) + \theta_i^a\Big) \end{aligned} \quad , \qquad (5)$$

the existence of a solution to any given PDA would be guaranteed upon proper quantization of the nonlinear functions $g(x)$ and $f(x)$. During learning, the sigmoid function $g(x)$ is used and $f(x)$ is defined as $f(x) = 2g(x) - 1$.

However, this proof does not exclude solutions with other neural net structures and does not necessarily guarantee the best learning behavior with third-order weights for all problems. In practice, second-order weights were used for some problems and good training results were achieved. The recurrent updating formula for second-order networks can be written as

$$\begin{aligned} S_i^{t+1} &= g\Big(\sum_{j,k} W^s_{ijk} S_j^t\,(\boldsymbol{R}^t \oplus \boldsymbol{I}^t)_k + \theta_i^s\Big) \\ A_i^{t+1} &= f\Big(\sum_{j,k} W^a_{ijk} S_j^t\,(\boldsymbol{R}^t \oplus \boldsymbol{I}^t)_k + \theta_i^a\Big) \end{aligned} \quad , \qquad (6)$$

where $(\boldsymbol{R}^t \oplus \boldsymbol{I}^t)_k$ is the concatenation of the two vectors $\boldsymbol{R}^t$ and $\boldsymbol{I}^t$, whose components are given by

$$(\boldsymbol{R}^t \oplus \boldsymbol{I}^t)_k = \begin{cases} R_k^t & \text{if } 0 < k \le N_R \\ I_{k-N_R}^t & \text{if } N_R < k \le N_I + N_R \end{cases} \quad . \qquad (7)$$

Experiments and comparisons between NNPDAs with different orders of connection weights were discussed in [Das92]. In most cases the third-order weights gave better learning results.

The existence proof of the NNPDA controller discussed above is based on the assumption of unary representations of internal states and symbols (both input and reading symbols). For the stack reading $\boldsymbol{R}^t$ and input $\boldsymbol{I}^t$, a unary representation (or linear independent vector representation) is necessary. This will be discussed in next subsection. However, unary representation of internal states may not be necessary. Moreover, to extract a discrete PDA, the procedure of state quantization is performed after learning and the quantized state vectors (often expressed in a binary form) are neither unary, nor linearly independent. But, during learning (especially hard problems), we often encounter the cases where we need to adjust independently the transitions between these linearly dependent state vectors. With third order weights the degrees of freedom are limited and each weight parameter does not associate with only one particular state transition as in the case of unary representations. Therefore, learning could be often trapped at a local minimum. To solve this problem, we propose a "full-order" connected network and find it very useful in learning some hard problems, like the Palindrome grammar. The basic formula of "full-order" network, for the example of one action output, is

$$A^{t+1} = f\left(\sum_{\{j\},k,l} W^{a}_{\{j\}kl}\left(S^{t}_{\{j\}} R^{t}_{k} I^{t}_{l}\right) + \theta^{a}\right) , \tag{8}$$

where the subscript $\{j\} \equiv \{j_1, j_2, ..., j_n\}$, represents all $2^n$ possible n-bit binary numbers ($j_m$=0, 1; m=1, 2, ..., n), and n is the number of state neurons. The state vector $\boldsymbol{S}^{t}_{\{j\}}$ is an $n_{th}$ order product of $\boldsymbol{S}^{t}$'s components defined as

$$S^{t}_{\{j\}} = \prod_{m=1}^{n} \left(j_m S^{t}_{m} + (1-j_m)(1-S^{t}_{m})\right) \quad . \tag{9}$$

In learning the palindrome grammar, the combination of Eq.(8) and the third order state dynamics of Eq.(5) were successfully used.

# 3.2 External Continuous Stack Memory

One of novel features of the NNPDA is the continuous stack memory. The continuous (or analog) stack was motivated by a desire to manipulate a stack with a gradient descent training algorithm. In order to minimize the error function along the gradient descent direction, the weight modification is proportional to the gradient of the error function

$$\Delta W \propto \frac{\partial}{\partial W}(ErrorFunction) \quad . \tag{10}$$

To couple the neural net with a stack memory, the stack variable must be included in the error function. One way of doing this is to make the stack variables a continuous function of the connection weights, so that an infinitesimal change of weights will cause an infinitesimal change of action values, which in turn cause an infinitesimal change of stack readings. Any discontinuity among these relations may cause the derivative to be infinity, thereby interfering with the learning process.

## 3.2.1 Continuous Stack Action

To fully describe the mechanism of the continuous stack, we discuss in detail: (1) the continuous stack action and stack operation; (2) how to read the stack and (3) the neural representation of the stack reading. Consider a conventional stack, as shown in Fig. 4(a), where there are stored a number of discrete symbols. The discrete stack actions include *pop*, *push* and *no-op*. Without affecting the generality of a stack function, it is assumed that each action only deals with one symbol. The *pop* simply removes the top symbol and the *push* places the symbol read from input string onto the top of stack. When the continuous stack is introduced, we have to replace both the discrete symbols in the stack by continuous symbols and the discrete *pop* and *push* actions by continuous actions. Therefore, we define the continuous *length* of every symbols. In Fig. 4(a), the stack is filled with discrete symbols and each symbol is interpreted as having equal length L=1. In the general case, as shown in Fig.4(b), the stack is filled with continuous symbols, each having a continuous length: $1 \geq L \geq 0$. These continuous symbols are generated by the continuous stack actions. As described in the neural network controller in Eqs.(5), (6) and (8), the output of the action neurons $A_i^t$ are calculated by the function $f(x)$ with analog values distributed within the interval [-1, 1]. The value of $A_i^t$ is interpreted as the intensity of the actions to be taken on the conventional stack [Harrison78]. When $A_i^t$ takes on continuous values, the natural generalization of the discrete dynamics is to interpret each continuous action $A_i^t$ as an uncertainty about the action to be taken. We represent this uncertainty in terms of the *length* of the discrete symbols to be pushed or popped. Therefore, at each time step only part of a discrete symbol is pushed or popped onto the stack with length determined by $|A_i^t|$. Whether to push or pop is determined by the sign of $A_i^t$: *push* if $A_i^t > \varepsilon$ and *pop* if $A_i^t < -\varepsilon$ where ε is a small number close to zero; otherwise a *no-operation* (*no-op*) takes place. After such actions, the stack construction would appear as in Fig.4 (b).

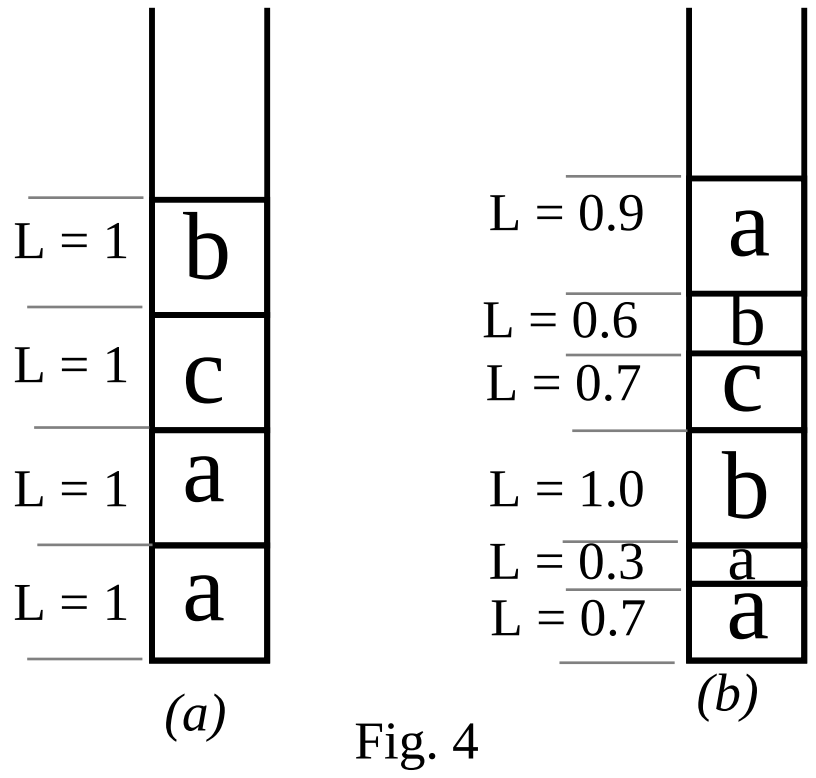

Fig. 4

Fig.4 Stack symbols with continuous lengths

(a) discrete stack is filled with discrete symbols which can be viewed as all having length = 1.

(b) continuous stack is filled with discrete symbols having continuous length: $0 \le L \le 1$.

In the above description of the stack operation, only one component of the vector $A_i^t$ is used and all three actions *pop*, *push* and *no-op* are represented by one variable. However, one could integrate continuous actions into a conventional discrete stack in many different ways. For instance, separate action neurons could be used to represent the different types of actions, i.e. one neuron with output $0 \le A_1^t \le 1$ to represent the value of *push* and another neuron with output $0 \le A_2^t \le 1$ to represent the value of *pop* action. In this case both $A_1^t$ and $A_2^t$ could simultaneously have nonzero output and the order in which the two actions (*push* and *pop*) are executed must be assigned in advance. If we first take a *pop* action and then *push*, we in effect introduce four types of actions in the discrete limit: (1) *push* ($A_1^t = 1$ and $A_2^t = 0$), (2) *pop* ($A_1^t = 0$ and $A_2^t = 1$), (3) *no action* ($A_1^t = 0$ and $A_2^t = 0$) and (4) *replace* ($A_1^t = 1$ and $A_2^t = 1$).

## 3.2.2 Reading the Stack

How to read from a continuous stack must be defined. For simplicity, we assume only one action neuron is used. In the conventional discrete stack a read operation only reads one symbol from the top of stack and sees nothing below. This reading method is not suitable for the continuous stack, since there will be a discontinuity in the content of the stack reading. More specifically, a reading discontinuity may happen in either of the following two cases: (1) after performing the action $A^t$, a symbol with very tiny length is left on the top of the stack; or (2) the top symbol has a very tiny (or zero) part being removed by the previous pop action $A^t$. In these two cases an infinitesimal perturbation to the action value $A^t$ could generate a discrete jump in the stack readings. See the example shown in Fig. 4(b). If $A^t = -0.9$, the symbol “a” will be popped entirely from the top of the stack. And the next reading $\boldsymbol{R}^{t+1}$ would be the symbol “b” with length = 0.6. However, if there is a small perturbation to the connection weights such that the value of $A^t$ increases by only 0.001, then $A^t$=-0.899. The top symbol “a” with length L=0.899 will be popped and a small portion of “a” remains on the top of stack. In that case the next reading $\boldsymbol{R}^{t+1}$ would be the symbol “a” with length = 0.001. A similar discrete jump will happen for the case where $A^t \approx 0$. To avoid this discontinuity we impose the condition that each time the continuous stack is read with depth equal to 1 from the stack’s top.

The advantages of this reading method are outlined below. First, a continuous reading function will be constructed with respect to the connection weights - any infinitesimal change of weights will cause an infinitesimal change of stack readings. In the example of Fig.4(b), for $A^t$=-0.9 the symbol “a” on the top is popped. The next reading contains two parts: symbol "b" with length = 0.6 and symbol "c" with length = 0.4 (the total length = 0.6 + 0.4 = 1.0). If the action value was changed to $A^t$=-0.899 due to a small perturbation of the connection weights, the symbol "a" is not totally popped off and a small fraction is left. In this case the next reading would contain: a small fraction of symbol "a" with length = 0.001, a part of symbol "b" with length = 0.6 and a part of symbol "c" with length = 0.399 (total length = 0.001 + 0.6 + 0.399 = 1.0). This example shows that the change of the next stack reading $\boldsymbol{R}^{t+1}$ is proportional to the change of previous action values $A^t$. When $\Delta A^t$ approaches zero, the change of readings $\Delta \boldsymbol{R}^{t+1}$ also approaches zero. It should be noted that this continuity of the reading function does not automatically guarantee that it is differentiable; and, even if it is differentiable, its derivative may not be a function feasible for numerical implementation. The complication of the derivatives $\partial \boldsymbol{R}^t / \partial \boldsymbol{W}$ and $\partial \boldsymbol{R}^t / \partial \boldsymbol{A}^\tau$ will be discussed in Appendix A.

The other advantage of the proposed reading method is its correspondence with a probabilistic interpretation of the continuous action value; a stochastic machine. The continuous action values can be interpreted as a type of uncertainty compared to the deterministic discrete *push* and *pop*. If the maximum of the absolute action value is one, *i.e.* $|A_i^t| \le 1$, the *length* of a symbol to be pushed or popped can be interpreted as the probability of this discrete action. Consequently, the reading of the stack with a total length equal to one implies the normalization of the total probabilities i.e. the summation of all the probabilities for reading each discrete symbol normalized to one. In other words, as in the previous example of Fig.4 (b), if the stack reading (with total length equals to one) contains: 'a' with length = 0.001, 'b' with length = 0.6 and "c" with length = 0.399, we can interpret that the stack symbol is being read with uncertainty: the probability of the read symbol to be "a" is very small as 0.001, the probability to be "b" is 0.6 and to be "c" is 0.399. When the stack length is less than 1, the reading may be only an 'a' with length = 0.1, this could be interpreted that the probability to read 'a' is 0.1 and the probability to read empty stack is 0.9.

### 3.2.3 Neural Representation

In the last subsections, the stack reading $\boldsymbol{R}^t$ and the input $\boldsymbol{I}^t$ are often described as a symbol. In this subsection, the actual neural representation of these two vectors will be discussed.

The neural representations of the input string symbol $I^t$ and the stack readings $R^t$ are determined under the following considerations. First, in the discrete limit (by quantization of the analog neurons to discrete levels) the learned neural network pushdown automata is required to behave the same way as a conventional pushdown automata. In this limit, since both sets $\{\boldsymbol{I}^t\}$ and $\{\boldsymbol{R}^t\}$ (each element of which corresponds to a symbol) represent the same set of discrete symbols, the neural representations of each $\boldsymbol{I}^t$ and $\boldsymbol{R}^t$ need to be identical. In this regard, there are no restrictions on their neural representations as long as they are the same. For instance, consider the symbols 'a', 'b' and 'e', the set $\{\boldsymbol{I}^t\}$ or $\{\boldsymbol{R}^t\}$ can be represented either by two neurons as (0, 1), (1, 0) and (1, 1) if a binary code is used or by three neurons as (1, 0, 0), (0, 1, 0) and (0, 0, 1) if an orthogonal code is used.

Second, during training, the stack reading should consist of continuous neuron values and each reading neuron $R^t$ should be able to represent the contents inside a segment of the continuous stack with total length = 1. This is in general a distributed mixture of the three possible symbols, each with a analog length less than 1. For effective neural information representation, it is important to require that there exist a unique one-to-one mapping between each vector $\boldsymbol{R}^t$ and the stack symbol component it represents.

The general mapping from the three continuous lengths to $\boldsymbol{R}^t$ can be written as

$$
\begin{gathered}
R^t = f(l_1, l_2, l_3, \vec{a}, \vec{b}, \vec{e}) \\
l_1 + l_2 + l_3 \le 1, \quad l_1 \ge 0, l_2 \ge 0, l_3 \ge 0
\end{gathered}
, \qquad (11)
$$

where $l_1$, $l_2$ and $l_3$ are the three continuous lengths of discrete symbols 'a', 'b', and 'e' contained in $R^t$ and $\vec{a}, \vec{b}, \vec{e}$ are the vector representations of 'a', 'b', and 'e' in neuron space. The condition $l_1+l_2+l_3 \le 1$(not $l_1+l_2+l_3 = 1$) includes the case of partial empty stack during training where the total length of symbols stored in the stack is less than one.

The first requirement for the discrete limit can be stated as

$$\begin{array}{lll} R^t = \vec{a} & if & l_1 = 1, l_2 = 0, l_3 = 0; \\ R^t = \vec{b} & if & l_2 = 0, l_2 = 1, l_3 = 0; \\ R^t = \vec{e} & if & l_3 = 0, l_2 = 0, l_3 = 1 \end{array} . \quad (12)$$

One simple way to satisfy this condition is to write $R^t$ as a linear combination of three basis vectors $\vec{a}, \vec{b}, \vec{e}$

$$R^t = l_1\vec{a} + l_2\vec{b} + l_3\vec{e} \ . \quad (13)$$

For the second requirement, uniqueness, the necessary and sufficient condition for the mapping in Eq.(13) is that the three neural vectors $\vec{a}, \vec{b}, \vec{e}$ be linearly independent. (By the uniqueness we mean that if there exists another set of coefficients $l'_1$, $l'_2$ and $l'_3$ such that $l'_1\vec{a} + l'_2\vec{b} + l'_3\vec{e} = l_1\vec{a} + l_2\vec{b} + l_3\vec{e}$ then $l'_1 = l_1$, $l'_2 = l_2$ and $l'_3 = l_3$.) If there are $m$ symbols used in the input strings, then at least $m$ analog neurons are needed to represent the input string symbol $I^t$ and the stack readings $R^t$ because any $m$ vectors in the lower, less than $m$, dimensional space would be linearly dependent on each other. In the three symbol example, this excludes the use of binary vectors (0, 1), (1, 0) and (1, 1) to represent symbols ‘a’, ‘b’ and ‘e’. For simplicity the unary neural representation, i.e. $\vec{a} = (1, 0, 0)$, $\vec{b} = (0, 1, 0)$ and $\vec{e} = (0, 0, 1)$ are used for the three symbols ‘a’, ‘b’ and ‘e’. In this case the stack readings $R^t$ are represented by a three-dimensional vector ($l_1$, $l_2$, $l_3$), indicating that in the current stack reading the lengths of letters ‘a’, ‘b’ and ‘e’ are $l_1$, $l_2$, $l_3$ respectively.

To conclude this section, a novel continuous stack is introduced. One interpretation of the *continuous stack* is the concept of *a magnitude associated with a discrete symbol*. This new concept stresses two aspects: (1) generalization of a discrete stack to a continuous stack and (2) identification of the stack readings and actions as neural network input and output with a probabilistic interpretation.

# 3.3 Dynamics of the Neural Network Pushdown Automata

For simplicity the following assumptions are made: (a) only deterministic pushdown automata are considered; (b) only one action neuron output $A^t$ is used; (c) the same set of symbols represent both the input and stack symbols, so that an action *push* only pushes the current input $\boldsymbol{I}^t$ onto the stack. These assumptions will restrict the class of CFG languages that the NNPDA can learn and recognize.

We illustrate the NNPDA dynamics by examples. Consider two symbol strings of ‘a’ and ‘b’. To mark the end of an input string the end symbol ‘e’ is introduced. A possible input string may be: “aababbabe.” Each time a string symbol ‘a’ (or ‘b’) is fed into the neural network controller, this same symbol ‘a’ (or ‘b’) could be pushed onto the stack (or the stack could be popped from the top) with magnitude $|A^t|$ according to the sign of $A^t$. The last symbol ‘e’ indicates the end of the input string. Upon receiving the end symbol, the neural network pushdown automata would generate a proper output to tell whether the input string was legal or illegal.

Numerically, two arrays are used to represent the stack: an integer array **stacksymbol**[] to store the symbols {‘a’, ‘b’, ‘e’} and a real number array **stacklength**[] for their lengths. A record of the number of symbols stored on the stack is kept in an integer **top**. Assume that four state neurons are used such that $S^t = (s_1, s_2, s_3, s_4)$, where $0 \le s_1, s_2, s_3, s_4 \le 1$ are the four neurons output.

The NNPDA operations are outlined for successive time steps.

(1) t = 0.

Initially, the stack is empty, so that **top** = 0 and the stack reading at $t = 0$ is $R^0 = (0, 0, 0)$. If the first symbol of the string is letter ‘a’, the initial input neural vector would be $I^0 = (1, 0, 0)$. Assume the initial state to be $S^0 = (1, 0, 0, 0)$. The stack is shown in Fig. 5(a).

(2) t = 1.

Initialize the NNPDA with the values $S^0$, $I^0$ and $R^0$ (as shown in Fig.3). After one iteration of Eq.(3), the new state

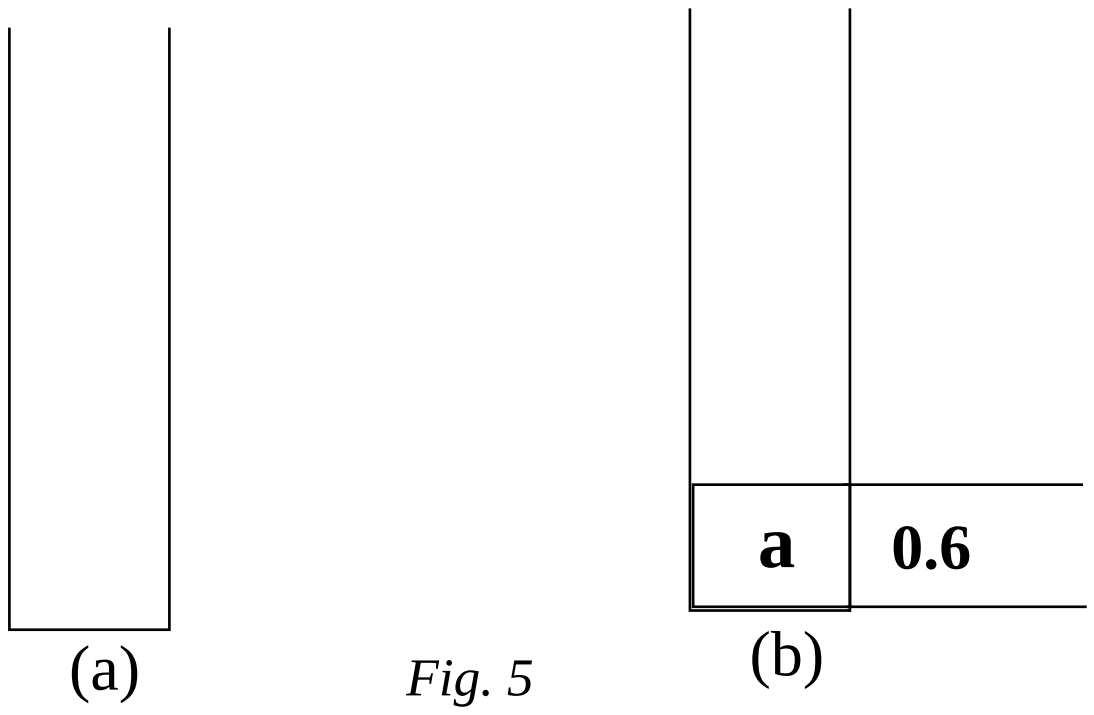

Fig.5 Stack status at (a) t = 0 and (b) t = 1.

$S^1$ and new action $A^1$ are obtained. Assume that the action output is $A^1 = 0.6$, then push symbol ‘a’ with length = 0.6 onto the stack. The new status of the stack can be represented as **stacksymbol**[1] =’a’, **stacklength**[1] = 0.6 and **top**=1. Then the next reading $R^1$ would be (.6, 0, 0). The stack is shown in Fig. 5(b).

If the next symbol in the input string is ‘b’, then $I^1 = (0, 1, 0)$. Substituting the new values $S^1$, $I^1$ and $R^1$ into Eq.(3) generates the next time values. Repeat the procedure.

(3) some later time t.

After several possible pushes, pops and no-ops, the current stack memory may have stored several continuous symbols as in Fig. 6(a): **top** = 4 (four symbols are stored), **stacksymbol**[] = (‘a’, ‘a’, ‘b’, ‘a’) and **stacklength**[] = (0.32, 0.2, 0.7, 0.4). Since the stack is read down from the top with depth = 1, the current stack reading would be $R^t = (0.4, 0.6, 0)$ as shown in Fig. 6(a). Assume the input symbol is ‘a’, so that $I^t = (1, 0, 0)$. The state vector can also be read from the state neuron output as $S^t$.

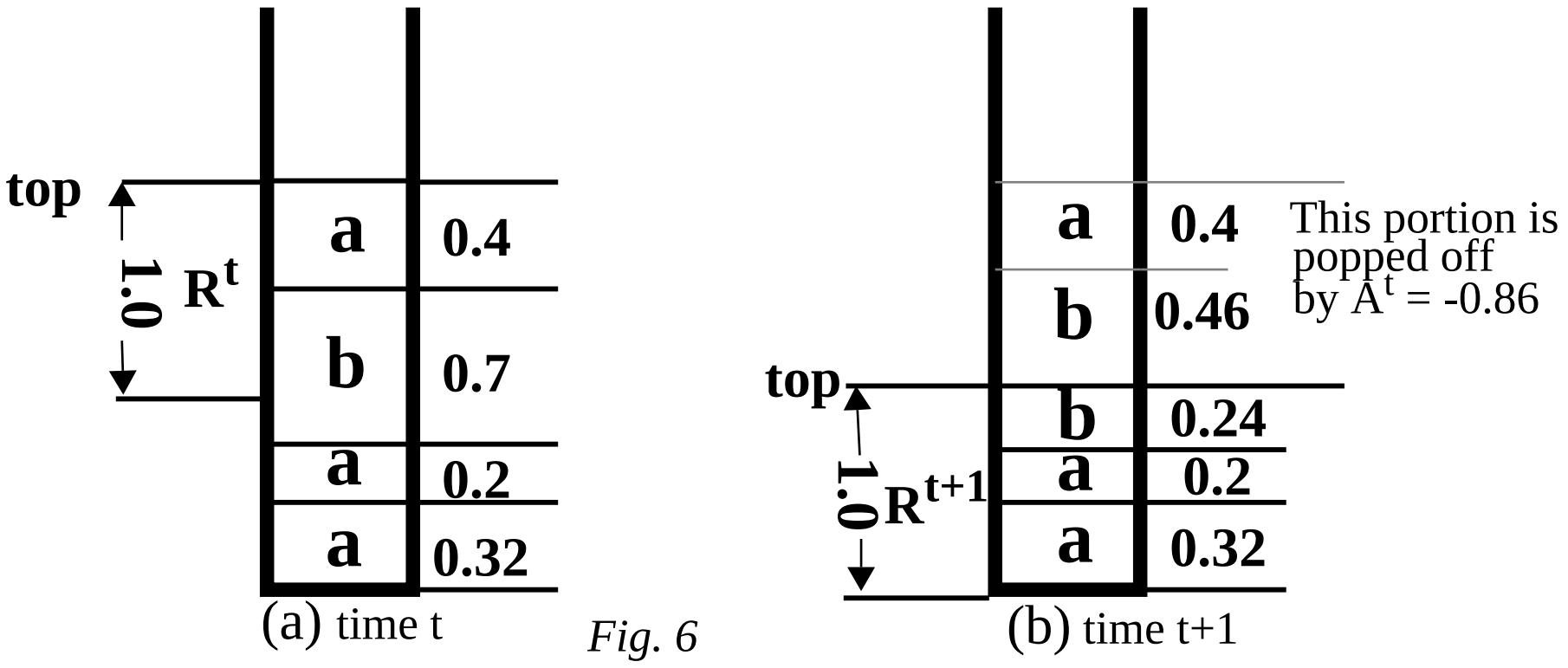

Fig.6 Continuous stack at (a) time t and (b) time t+1.

(4) time t+1.

Substitute $S^t$, $I^t$ and $R^t$ into Eq.(3) and the next time values are obtained. If the action $A^{t+1} = -.86$, a segment of the stack with content of length = 0.86 is popped. This “popped segment” includes 0.4 of ‘a’ and 0.46 of ‘b’ and the stack now has **top** = 3 (three symbols are left), **stacksymbol**[] = (‘a’, ‘a’, ‘b’) and **stacklength**[] = (0.32, 0.2, 0.24). The next stack reading would be $R^{t+1} = (.52, .24, 0)$ (formed by 0.32 of ‘a’ plus 0.2 of ‘a’ plus 0.24 of ‘b’).

This procedure is repeated until the end of the input string. The classification of an input string is determined by examining the final state neuron output and the stack length. The criterion for training and classification will be discussed in the next two sections.

## 3.4 Objective Function

The objective function to be minimized is defined as a scalar error measure which is a function of both the end state and the stack length. For a conventional pushdown automata, either the end state or the stack length alone is a sufficient criterion to determine the acceptance of input strings [Harrison78]. If either the end state reaches a desired final state, or the stack is ended empty, the input string is legal; otherwise illegal. However in training the NNPDA we find that a combination of the two criteria seems necessary. We speculate that this is because of the existence of too many local minimum in phase space. Thus, an objective function consisting of only one criteria of final state or stack length will have a very complex phase space configuration so that the local learning algorithm - gradient descent - would not be able to drive the system from the local minima. Therefore, a legal string is required to satisfy both conditions: (1) at the end the NNPDA reaches a desired final state *and* (2) the stack is empty.

Define the stack length at time $t$ to be $L^t$. Then, $L^t$ can be evaluated recursively in terms of the action value $A^t$

$$L^{t+1} = L^t + A^t \quad , \tag{14}$$

because only the push or pop actions can change the length of stack. The initial condition is $L^t = 0$ and the constraint $L^t \geq 0$ should be imposed at all the times. Let $T$-1 be the final time at the end of input string. For legal strings the straightforward error function $E$ to be minimized could be

$$E = (\boldsymbol{S}_f - \boldsymbol{S}^T)^2 + (L^T)^2 \quad , \tag{15}$$

where $\boldsymbol{S}_f$ is the desired final state. However, this error function could not be used to train illegal strings. For illegal strings the desired value of function $E$ is not known. Maximizing the same error $E$ as in Eq.(15), in general, would not give a correct answer because $E$ is an unbounded function and an illegal string may not end with a long stack length. However, replacing $\boldsymbol{S}_f$ in Eq.(15) with a desired end state for illegal strings and then minimizing $E$ presents the same problem since illegal strings are required to end with an empty stack (in effect avoid using stack). The main difficulty is that there is not enough information to decide the desired value of stack length for illegal strings.

In general, the following reasoning is applied. Since a legal string requires **both** (a) the desired final state $\boldsymbol{S}^T\!=\!\boldsymbol{S}_f$, **and** (b) an empty stack ($L^t = 0$); an illegal string should require the opposite: **either** (a) the final state be a large measurable distance from $\boldsymbol{S}_f$, **or** (b) a non-empty stack ($L^t \geq 1$). Although other training requirements could be defined, in practice, both of these conditions are successfully used.

One way to implement the above requirement is to introduce a unified error function $E$ which can be used to train both legal and illegal strings. For simplicity we assign the final state(s) in such a way that only one neuron $S_{Ns}$ output is to be checked at time $T$ at the end of input string. We require ${S_{Ns}}^T = 1$ *and* $L^T = 0$ for legal strings and ${S_{Ns}}^T = 0$ *or* $L^T \geq 1$ for illegal strings. In this case the unified error function to be minimized for both legal and illegal strings can be defined as

$$E = (v + L^T - S^T_{N_S})^2 \equiv e^2 \quad , \tag{16}$$

where $v$ is a parameter assigned as a target value for each training example. For legal strings $v = 1$ and for illegal strings $v = \min\{0, {S_{Ns}}^T\text{-}L^T\}$. The learning algorithm is derived by minimizing this error function with the proper value of $v$ for each input string. Correctness of the error function(16) can be checked separately for each string. If the input string is legal, $v = 1$. Then, minimizing $E$ corresponds to the requirement that ${S_{Ns}}^T\!=\!1$ *and* $L^T\!=\!0$ — the desired final state and empty stack. If the input string is illegal, we require $v = \min\{0, {S_{Ns}}^T\text{-}L^T\}$. There are two possible cases. First, when ${S_{Ns}}^T > L^T$, let $v = 0$, which implies that minimizing $E$ corresponds to driving $L^T$ to approach ${S_{Ns}}^T$. The minimum of $E$ can be reached if ${S_{Ns}}^T = L^T$. This means that for each input string (neuron activity ${S_{Ns}}^T$ is discretized to 0 or 1) one of the following requirements is met: ${S_{Ns}}^T = 0$ *or* $L^T = 1$. Second, if $L^T$ is already greater than ${S_{Ns}}^T$, then $v = \min\{0, {S_{Ns}}^T\text{-}L^T\} = {S_{Ns}}^T\text{-}L^T$. This leads to $E$=0, implying “do not care” or “no error”. Thus, in the discrete limit, the combination of the two cases corresponds a requirement for illegal strings: either ${S_{Ns}}^T = 0$ (illegal state) *or* $L^T \geq 1$(non-empty stack).

From the above analysis for analog values of $S_{Ns}{}^{T}$, the expression $H \equiv S_{Ns}{}^{T} - L^{T}$ could be considered as a continuous measure of how well both of the two conditions $S_{Ns}{}^{T} = 1$ *and* $L^{T} = 0$ are satisfied. The desired value for legal string is $H$=1 and for illegal strings $H \leq 0$. This $H$ function also provides a simple test measure for new input string strings. After training we will use the same measure $H \equiv S_{Ns}{}^{T} - L^{T}$ to test the generalization capability of the NNPDA on unseen input strings. The measure $H$ will be evaluated for each input string. A string is classified as legal if $H > .5$, otherwise illegal.

Another criterion to assist learning is the "trap state," one of the "hints" used by [Das93]. This "trap state" is used in training the non-trivial Palindrome grammar; details are discussed in Section IV.

# 3.5 Training Algorithm

The training algorithm is derived by minimizing the error function using a gradient descent optimization method. There are currently two ways to implement gradient descent optimization in recurrent neural networks: the chain-rule differentiation can be propagated forward or backward in time. The forward propagation method is also known as Real Time Recurrent Learning (RTRL) [Williams89], which propagates a sensitivity matrix forward in time until the end of an input sequence. Then, error correction is performed and the weights are modified according to the error message and the sensitivity matrix. Back-propagation-through-time [Rumelhart86b] can be applied to recurrent network training by unfolding the time sequence of mappings into a multilayer feed-forward net, each layer with identical weights. This method requires memorizing the state history of input sequence and, whenever the error is found, the error must be propagated backward in time to the starting point. Due to the nature of the backward path, it is an off-line method. In principle, both methods can be generalized to couple the external stack memory with recurrent neural network and train the NNPDA. RTRL is desirable for on-line training because the weights can be modified immediately after the error is detected without waiting for back-propagation. But it has a complexity of $O(N^4)$ compared to the complexity of $O(N^3)$ for back-propagation through time (N is the number of neurons and first order connection weights are assumed). For the task of grammatical inference, on-line training is not necessary because error messages are only given at the end of input strings. But, since the derivation of forward propagation algorithm is more straightforward for NNPDA, we first consider the generalization of RTRL for training the NNPDA.

From Eqs.(10) and (16), the weight correction for gradient descent learning becomes

$$\Delta W = -\eta \left(v + L^{T} - S^{T}_{N_S}\right) \left(\frac{\partial L^{T}}{\partial W} - \frac{\partial S^{T}_{N_S}}{\partial W}\right) \quad , \tag{17}$$

where $\eta$ is the learning rate and the partial derivatives of $L^{T}$ and $S^{T}_{Ns}$ with respect to weight matrix $W$ can be calculated recursively. The formula for $\partial L^{t} / \partial \boldsymbol{W}$ is easily derived from Eq.(14)

$$\frac{\partial L^{t+1}}{\partial W} = \frac{\partial L^{t}}{\partial W} + \frac{\partial A^{t}}{\partial W} \quad . \tag{18}$$

The recursions for $\partial \boldsymbol{S}^{t} / \partial \boldsymbol{W}$ and $\partial A^{t} / \partial \boldsymbol{W}$ are found by differentiating the controller dynamical equations. For example the second-order connection weights of Eq.(5) yield

$$\frac{\partial S^{t+1}_{i'}}{\partial W_{ijk}} = h_{i'}(S^{t}_{i'}) \left( \delta_{ii'} S^{t}_{j} (\boldsymbol{R}^{t} \oplus \boldsymbol{I}^{t})_{k} + \sum_{j'=1}^{N_S} \sum_{k'=1}^{2N_I} W_{i'j'k'} (\boldsymbol{R}^{t} \oplus \boldsymbol{I}^{t})_{k'} \frac{\partial S^{t}_{j'}}{\partial W_{ijk}} + \sum_{j'=1}^{N_S} \sum_{k'=1}^{N_I} W_{i'j'k'} S^{t}_{j'} \frac{\partial R^{t}_{k'}}{\partial W_{ijk}} \right) \quad . \tag{19}$$

It should be noticed that Eq.(19) is an abbreviation of four equations for $\partial S^{t+1}{}_{i'} / \partial W^{s}{}_{ijk}$, $\partial S^{t+1}{}_{i'} / \partial W^{a}{}_{jk}$, $\partial A^{t+1} / \partial W^{s}{}_{ijk}$ and $\partial A^{t+1} / \partial W^{a}{}_{jk}$. For simplicity the notations of $S^{t}$ and $A^{t}$ are combined into one equation. The $(N_S+1)_{th}$ component of vector $\boldsymbol{S}^{t}$ is $A^{t}$. The function $h_i(x)$ represents derivatives $g'(x)$ for i =1 to $N_S$ and $f'(x)$ for i = $N_{S+1}$. $W^{s}$ and $W^{a}$ are similarly combined such that $W_{ijk}$ represents $W_{ijk}{}^{s}$ for i=1 to $N_S$ and $W_{jk}{}^{a}$ for i=$N_S$+1. (Note the assumption that $N_A$=1 and $N_R$=$N_I$). The learning algorithm formulas for the third order state transition and "full order" action mapping are presented in Appendix B.

From these recursions and knowing the initial conditions of $\partial \boldsymbol{S}^{0} / \partial \boldsymbol{W}$, $\partial A^{0} / \partial \boldsymbol{W}$, their values at a later time can be evaluated by Eq.(19). But, the recursion is not complete until $\partial \boldsymbol{R}^{t+1}/\partial \boldsymbol{W}$ is expressed in terms of $\partial \boldsymbol{S}^t/\partial \boldsymbol{W}$, $\partial A^t/\partial \boldsymbol{W}$ and $\partial \boldsymbol{R}^t/\partial \boldsymbol{W}$. This relation may not be easy to find, since the stack reading is a highly nonlinear function of all the previous actions and input symbols, as shown in Eq.(4), $\boldsymbol{R}^t=F(\boldsymbol{A}^1, \boldsymbol{A}^2, ..., \boldsymbol{A}^t; \boldsymbol{I}^1, \boldsymbol{I}^2, ..., \boldsymbol{I}^t)$. The approximate recursive relation for $\partial \boldsymbol{R}^{t+1}/\partial \boldsymbol{W}$ can be derived (for details see Appendix A). To the lowest order in its expansion, we have

$$\frac{\partial R^t_{k'}}{\partial W_{ijk}} \approx (\delta_{k' r^t_1} - \delta_{k' r^t_2}) \frac{\partial A^t}{\partial W_{ijk}} , \tag{20}$$

where $r_1{}^t$ and $r_2{}^t$ are the ordinal numbers of neurons that represent the top and the bottom symbols respectively in the reading $R^t$. Consider for example the case where after the execution of the action $A^t$, the stack is (from bottom to top): (0, 0.9, 0), (.2, 0, 0), (0, .7, 0) and (0, 0, .15). Then $r_1{}^t$=3 and $r_2{}^t$=1, because the symbol (0, 0, .15) on the top is the third symbol and the symbol (.2, 0, 0) on the bottom of $R^t$ is the first one.

The complete recursive equations Eqs.(18), (19) and (20), together with the NNPDA dynamical equations can be forward propagated with initial conditions $\partial \boldsymbol{S}^0/\partial \boldsymbol{W}=0$, $\partial A^0/\partial \boldsymbol{W}=0$ and $\partial \boldsymbol{R}^0/\partial \boldsymbol{W}=0$. The initial values of $A^0$ and $\boldsymbol{R}^0$ are zero and the initial state $\boldsymbol{S}^0$ could be assigned any constant. At the end of the input string, the weight correction Eq.(17) is evaluated. The final weight correction can be performed using either batch or stochastic learning.

However, there is the case of “pop empty stack.” If the total length of the remaining symbols in the stack is less than the value of a pop action ($L^{t-1}<|A^t|$), a “*pop empty stack*” occurs. For a well designed conventional pushdown automata “*pop empty stack*” never occurs. But, in learning a PDA, whether with a NNPDA or another method, such an action seems almost inevitable. We devise two possible ways to deal with this case. First, the input sequence can be interrupted whenever a “pop empty stack” occurs and weight corrections are made to increase the stack length ($\Delta W \sim \partial L^t/\partial W$). And, second, when we have “pop empty stack” and the input string is illegal, no weight correction is made. Conversely, weight corrections are made for legal input strings.

## 3.6 Extraction of PDA from a Trained NNPDA

After training with examples of a context free grammar, the NNPDA in general could recognize correctly the training set up to a certain length of strings. But, because of the analog nature of NNPDA, the recognition results are not “correct” in the discrete sense. The final state output are analog values between 0 and 1, which are usually reduced to the binary values of 0 and 1 by a threshold of 0.5. Thus, analog recognition errors still exist and could accumulate as the input strings become longer. To extract from the trained NNPDA a PDA which represents the underlying CFG, we devise a quantization procedure that converts an analog NNPDA to a discrete PDA. To simplify the state structure of the extracted discrete PDA, a minimization procedure for the PDA must be devised.

The quantization can be performed as follows. First, the action neuron(s) is quantized into three discrete values: -1, 0 and 1 according to the rule

$$A = \begin{cases} 0, & \text{if} \quad (|A| \le A^*) \\ -1, & \text{if} \quad (A < -A^*) \\ 1, & \text{if} \quad (A > A^*) \end{cases} , \tag{21}$$

where the threshold $A^*$ was chosen to be 0.5 for most of our numerical simulations (However, our experience indicates that the quantization results do not seem sensitive to the selection of $A^*$ values and other values besides 0.5 could be used). In this way the continuous stack will behave like a discrete stack and generate the discrete actions: push, no-op and pop actions. Next we perform a cluster analysis of the internal states. All input strings that have been recognized correctly are fed into the trained NNPDA and a set of analog internal states is generated. This set is divided into several clusters using a standard *K*-mean clustering algorithm [Duda73]. The number of clusters *K* is determined by minimizing the averaged distance from each state to its cluster center (in case the clusters are not well separated more training with these strings may be needed). After the cluster analysis store the cluster centers as the representative points of quantized internal states, then a PDA with discrete states is created and the number of states is equal to the number of clusters. During further testing, each analog internal state is quantized to its nearest cluster representative points and the discrete transition rules can be extracted. Now construct a transition diagram and this is the extracted PDA.

In some cases, instead of quantizing the whole state vectors, quantizing each of the state neurons is also useful. If the state neuron's output is distributed near their saturation values (0 or 1), a binary quantization is natural, i.e. $S^t_i$ is quantized to one if $S^t_i > 0.5$ and zero otherwise. If the state neural activity is uniformly distributed, more quantization levels are needed. The quantized NNPDA is tested with training or test strings again. If the recognition is incorrect, a finer re-quantization is needed (see [Giles92a] for a discussion of a similar method for FSA extraction for trained NN-FSA).

When a linear "full order" mapping is used for the action output (linear "full order" mapping is the linear form of Eq.(8)), then the quantization rule of Eq.(21) can be replaced by quantizing the connection weights by:

$$W^a = \begin{cases} 0, & \text{if } (|W^a| \le W^*) \\ -1, & \text{if } (W^a < -W^*) \\ 1, & \text{if } (W^a > W^*) \end{cases}, \tag{22}$$

where $W^a$ are the connection weights for action output and $W^*$ is the threshold. For details, see the numerical simulation for learning the Palindrome grammar.

After extraction of the discrete PDA, we reduce the state structure by pruning equivalent states. It is known that, in general, there exists no minimization algorithm (as for FSAs) for obtaining the unique minimal PDA; and that there exists no algorithm to tell whether or not two context free grammars or the two PDAs which accept two context free grammars are equivalent [Hopcroft79]. But, for a given specific structure of a PDA, the minimal size can be obtained by exhaustive search. For instance, assume a specific structure of a deterministic PDA, which pushes and pops only one symbol per input and the stack symbols are the same as input symbols. For this type of PDA each state transition can be characterized by a three-tuple condition $(\alpha,\beta,\gamma)$, where $\alpha$ is input symbol, $\beta$ is stack reading symbol and $\gamma$=1, -1, 0 represents push, pop and no-op. If we consider each combination of $(\alpha,\beta,\gamma)$ as an equivalent input symbol of a regular grammar, the extracted PDA transition diagram is equivalent to a finite state automaton transition diagram where a transition occurs each time a "symbol" $(\alpha,\beta,\gamma)$ is seen. Thus, the minimization algorithm for FSA can also be effectively used to reduce the extracted PDA. For detailed examples, see the next section.

# IV. NUMERICAL SIMULATIONS (learning grammars)

To illustrate the learning capabilities of the NNPDA, we train the NNPDA on a finite number of positive and negative strings of three context-free grammars. Different types of NNPDA and training procedures are discussed for each particular problem set. For all problems the external stack of the NNPDA is initially empty. All simulations were performed with 64 bit, double precision. For training we started with short strings and gradually increased the string length [Elman91]. For some simulations only 5 significant figures are presented.

## 4.1 Balanced Parenthesis Grammar

We train a second-order NNPDA to correctly recognize a given sequence of "balanced" parentheses. Input sequences consist of two input symbols '(' and ')' and an end symbol 'e'. Unary input representations are used with three input neurons, where (1,0,0), (0,1,0) and (0,0,1) represent respectively '(', ')', and 'e'. The stack action is controlled by one action neuron $A^t$. The number of state neurons is chosen empirically to be three, since the correct PDA controller is known to be a two state machine. The initial state is (1, 0, 0). At the end of the input string the value of third state neuron $S_3$ is checked. During training, the target value of $S_3$ is 1.0 for legal strings and 0.0 for illegal strings.

The training set consists of fifty strings: all thirty possible strings up to length four and twenty randomly selected longer strings up to length eight. The training criterion and algorithm (RTRL) are the same as described in Sections 3.4-5. For each run the initial weights are randomly chosen from the interval [-1,1]. For 5 different runs approximately one hundred training epochs are needed for the NNPDA to converge, i.e. learn the entire training set. To speed up training, we introduce the empirical condition that the input sequence is stopped and the stack length is reduced ($\Delta \boldsymbol{W} \sim -\partial L^t/\partial \boldsymbol{W}$) if a "*pop empty stack*" occurs during input of an illegal string. In this case, after only twenty epochs of training, the training set is learned. During testing, all the strings up to length twenty can be correctly recognized (totally $2^{21}$ strings). The acceptance criterion is discussed in Section 3.4. Due to analog error accumulation, longer strings could not be correctly recognized. To extract a discrete PDA the state neuron activation [0, 1] is quantized into five segments:

(0, 0.125), (0.125, 0.375), (0.375, 0.625), (0.625, 0.875), (0.8755, 1) or five discrete values: $S_i$= 0, 0.25, 0.5, 0.75 and 1.0, each corresponding to one segment. After quantization, the analog NNPDA becomes a discrete PDA. To check its performance, randomly chosen longer strings (length 50 to 100) were tested. All strings incorrectly classified by the analog NNPDA were now correctly recognized by the discrete PDA.

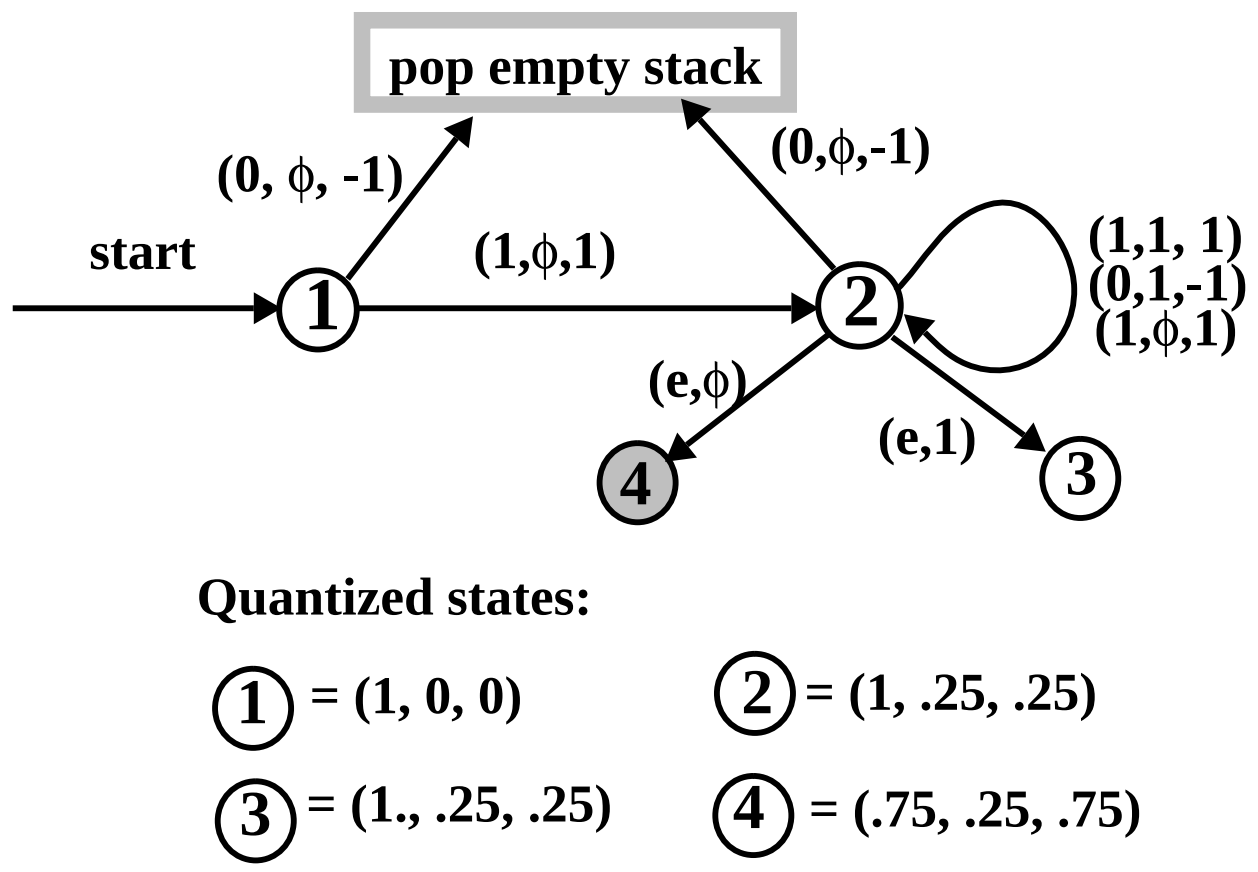

Fig. 7

Fig.7 The pushdown automaton (PDA) extracted from the NNPDA after the balanced parenthesis grammar was learned. The discrete states (1), (2), (3) and (4) are obtained by quantizing the numerical values of state neurons into five levels: 0, .25, .5, .75, and 1. State (1) is start state. State (4) is the legal end state. Just before the end symbol, a legal string must end at state (2) with an empty stack.

The transition diagram is extracted by tracing all possible paths of state transition numerically. This is easily done using a tree search method. Denote each node of the tree as a combination of state and stack reading. Starting from the root node, the initial state and empty stack, input all possible symbols at each node and trace the path of each symbol by calculating the next time state, stack reading and stack operation in terms of quantized NNPDA. Each time a new node is calculated, this node is checked to see if it has already been created in the previous level of the tree. If it is not, create this node and construct a transition line from the old node to the new node. Label the stack operation for this transition. Repeat this procedure at the new node until no additional new node occurs. The result of the tree structure can be translated to a transition diagram with each state as a node. As shown in Fig.7, each circle represents one quantized neural state and the arrows represent the state transitions. The notation (a,b,c) in Fig.7 represents a transition that occurs when the input symbol is $\boldsymbol{I}^t$= ‘a’, the stack reading $\boldsymbol{R}^t$= ‘b’ and action neuron output is $\boldsymbol{A}^t$=c. The two parentheses ‘(’ and ‘)’ are denoted by ‘1’ and ‘0’ and an empty stack reading by ‘ϕ’. It is seen from Fig.7 that when a ‘1’ is presented to the NNPDA, a ‘1’ is pushed onto the stack (due to rules (1,ϕ,1) and (1,1,1)). If a ‘0’ is presented to the NNPDA, a ‘1’ is popped from the stack (due to (0,1,-1)). Whenever a ‘0’ is presented and the stack is empty, the “pop empty stack” occurs. An input string will be classified as legal if, just before the presentation of the end symbol, the PDA is at state 2 and the stack is empty. Otherwise the input string is illegal. i.e. either “pop empty stack” occurs or the stack is not empty). *This is indeed the desired PDA.* In addition to the start state (state 1), only one state (state 2) is needed. States 3 and 4 are only needed to check if the stack is empty at the end of string.

## 4.2 $\mathbf{1^n0^n}$ grammar.

The language of the $\mathbf{1^n0^n}$ grammar is a subset of the parenthesis grammar. The $\mathbf{1^n0^n}$ PDA needs at least 2 internal states in order to filter out the strings legal for the balanced parenthesis grammar but illegal for the $\mathbf{1^n0^n}$ grammar [Hopcroft79]. The neural controller we used to learn the $\mathbf{1^n0^n}$ grammar had 5 state neurons.

A small training set, 27 short strings with 12 legal and 15 illegal strings shown below was initially used for training:

| | | | | |
|---|---|---|---|---|
| n1 | n11 | n1000 | y1100 | n1011 |
| y10 | y10 | y1100 | n110010 | y10 |
| n0 | n100 | n1111 | y11110000 | n1101, |
| y10 | y10 | y1100 | n110100 | |
| n00 | n1001 | n1110 | y1111100000 | |
| y10 | y1100 | n101100 | n1010 | |

where the letter 'n' and 'y' in front of the strings denote the classifications "no" and "yes". The $\mathbf{1^n0^n}$ grammar contains very few legal strings; among $2^L$ strings of length L there is only one legal string 11...100...0. Hence, the training set replicates some of the short legal strings "10" and "1100" between illegal strings in order to give balanced training set. For this example, the empirical rules (or "hint") of "pop empty stack" or "dead state" are not used. Whenever a negative stack length appears, we stop and modify the weights to increase the stack length $L^t$ ($\Delta \mathbf{W} \sim \partial L^t / \partial \mathbf{W}$). This is equivalent to increasing the "*push*" action value $A^t$ to avoid "*pop* empty stack".

After 100 training epochs, the NNPDA correctly classified the training set and was tested on unseen strings. Up to length eight, all strings are classified correctly except the following six strings:

n11000 n1100100 n01110000 n10101000 n11011000 n11001100.

These strings are then added to the training set and the NNPDA is retrained for another 100 epochs. Testing found 8 errors for all strings up to length nine. The misclassified strings are again added to the training set. After repeating this procedure five times, the trained NNPDA correctly classified all 2,097,150 strings up to length twenty and 20 randomly chosen strings up to length 160.

To analyze the learned NNPDA, the state neurons are quantized into two levels: 0 (if $S_i$<0.5) and 1 (otherwise), and the action neuron is quantized into three levels: -1, 0 and 1 as before. Starting from the initial state (1,0,0,0,0) and empty stack, all possible state transitions could be identified by inputting different strings. The resultant transition diagram is shown in Fig.8, where six binary states: (1,0,0,0,0), (1,0,0,0,1), (0,0,0,0,1), (1,1,1,1,1), (0,0,0,1,1) and (1,0,1,1,1) were found to form a close loop for any input strings of '0' and '1'. For clarity, the transitions for inputting an end symbol are not shown. Without end symbol, the state (1,1,1,1,1) is the desired final state for legal strings. All other states are illegal final states. This is because that starting from (1,1,1,1,1) with an empty stack, an end symbol input will lead to state (0,0,0,0,1). But, in all other cases (either starting from state (1,1,1,1,1) with non-empty stack or starting from other states) an end symbol input will lead to an illegal final state (*,*,*,*,0), a state with last neuron activity being zero.

The state transition diagram of the extracted PDA can be reduced using procedures previously discussed. The reduced transition diagram is shown in Fig.9, where the states 1, 2, 3 and 4 represent the quantized states (1,0,0,0,0), (1,0,0,0,1), the combination of states {(1,1,1,1,1), (1,0,1,1,1)} and the combination {(0,0,0,0,1), (0,0,0,1,1)} respectively. In the reduced diagram, state 3 is the desired final state. Recall that acceptance of a legal string requires both a desired final state and an empty stack.

## 4.3 Palindrome grammar.

The language of the deterministic Palindrome grammar contains all strings in the form of $WcW'$, where $W$ represents an arbitrary string of given symbols (here, we use two symbols '*a*' and '*b*'), $W'$ is the reversed order of $W$, and '*c*' is an additional symbol to mark the boundary symbol between $W$ and $W'$. For example, strings "abaaabbcbbaaaba" or "bbabbacabbabb" are legal.

The minimal (to our knowledge) palindrome PDA is shown in Fig. 10. Starting with state (1), every input symbol 'a' or 'b' is pushed onto the stack and the PDA remains in state (1). After an input symbol 'c' the PDA moves to state (2). When in state (2) the PDA pops every stack symbols if the stack reading ('a' or 'b') matches the input symbol; otherwise it moves to a trap state. The input string is classified as legal only if the PDA ends at state (2) with empty stack. In this example no end symbol is used.

This grammar has been found difficult to learn [Das92]. In our numerical simulations, both second order and third order nets were not able to learn a correct PDA for palindrome grammar. Two major difficulties were found. First, we lack sufficient information to supervise the stack actions for illegal strings. In most simulations the NNPDA did not learn to push correctly every symbol into the stack for illegal strings like "ab" and "babbaa" since it was not told what should be the target stack length during training. After seeing 'c' as in strings "abcba" (legal) or "babaacaab" and "ba-baacabb" (both illegal), the NNPDA is supposed to compare input symbols with stack readings and perform a pop if they match. But, since those symbols before 'c' were not stored in the stack as discrete symbols, the NNPDA could not compare the right stack symbols with input and perform the correct pops. Although, in learning the balanced parenthesis grammar a NNPDA had been able to learn a correct pop, this is a different level of stack operation. Comparing the two transition diagrams in Figs.7 and 10, it can be found that the palindrome grammar involves a more sophisticated level of stack manipulations than those in the balanced parenthesis grammar PDA. The stack of balanced parenthesis grammar is in fact only a counter. As shown in Fig.7, all the state transitions and stack actions can be de-cided totally by the combination of input symbol and current state, they do not really depend on the contents the stack is reading. (In this sense, only a second order correlation is needed.) But, the stack actions for the palindrome grammar require a third order correlation and actual dependence on the stack contents.

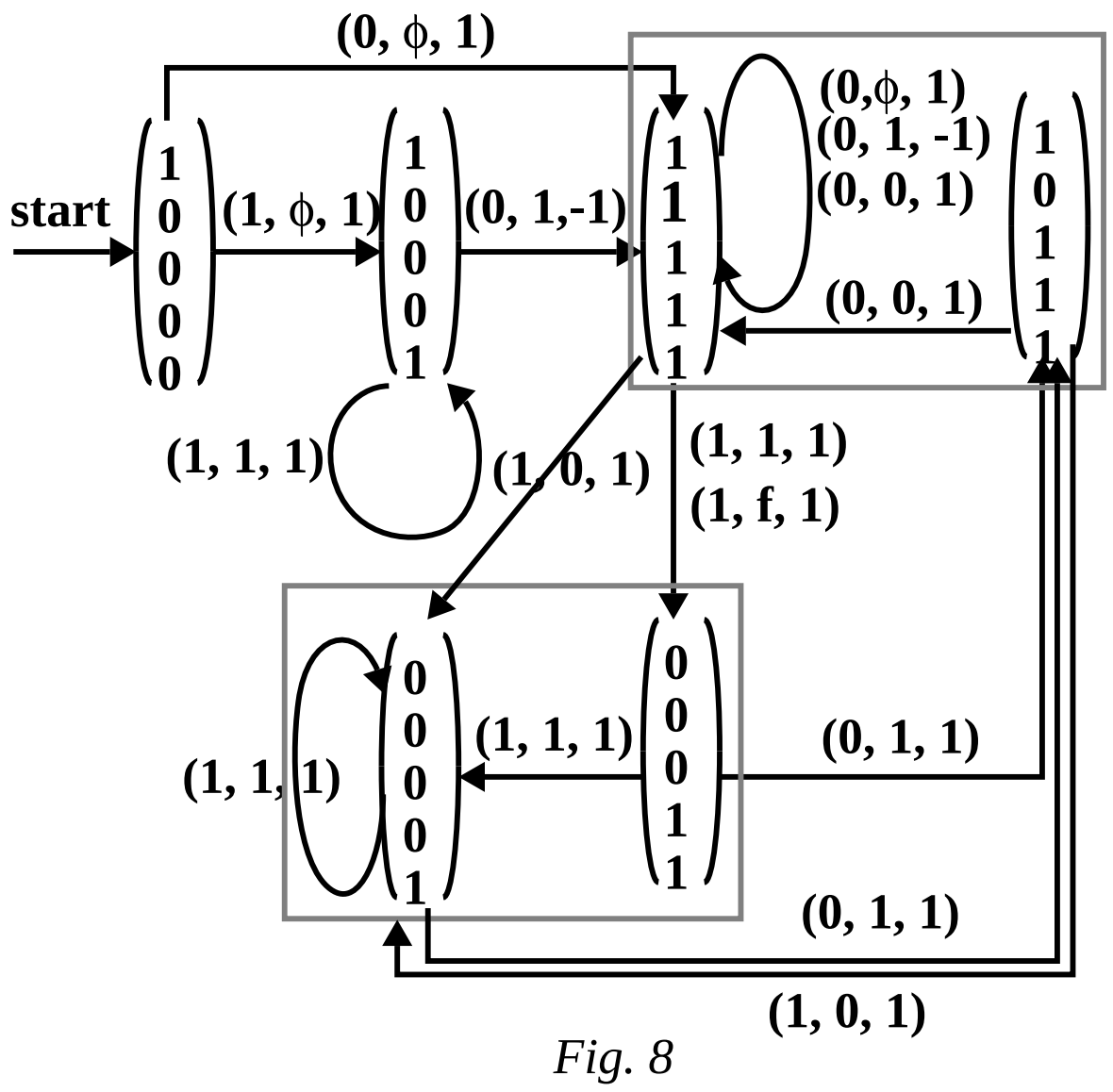

*Fig. 8*

Fig.8 The state transition diagram extracted from the trained NNPDA where the training examples were from the context-free grammar $1^n0^n$. In the figure, each five-component column vector represents a state of the PDA which is obtained by quantizing each of the state neurons to the binary values: 0 and 1.

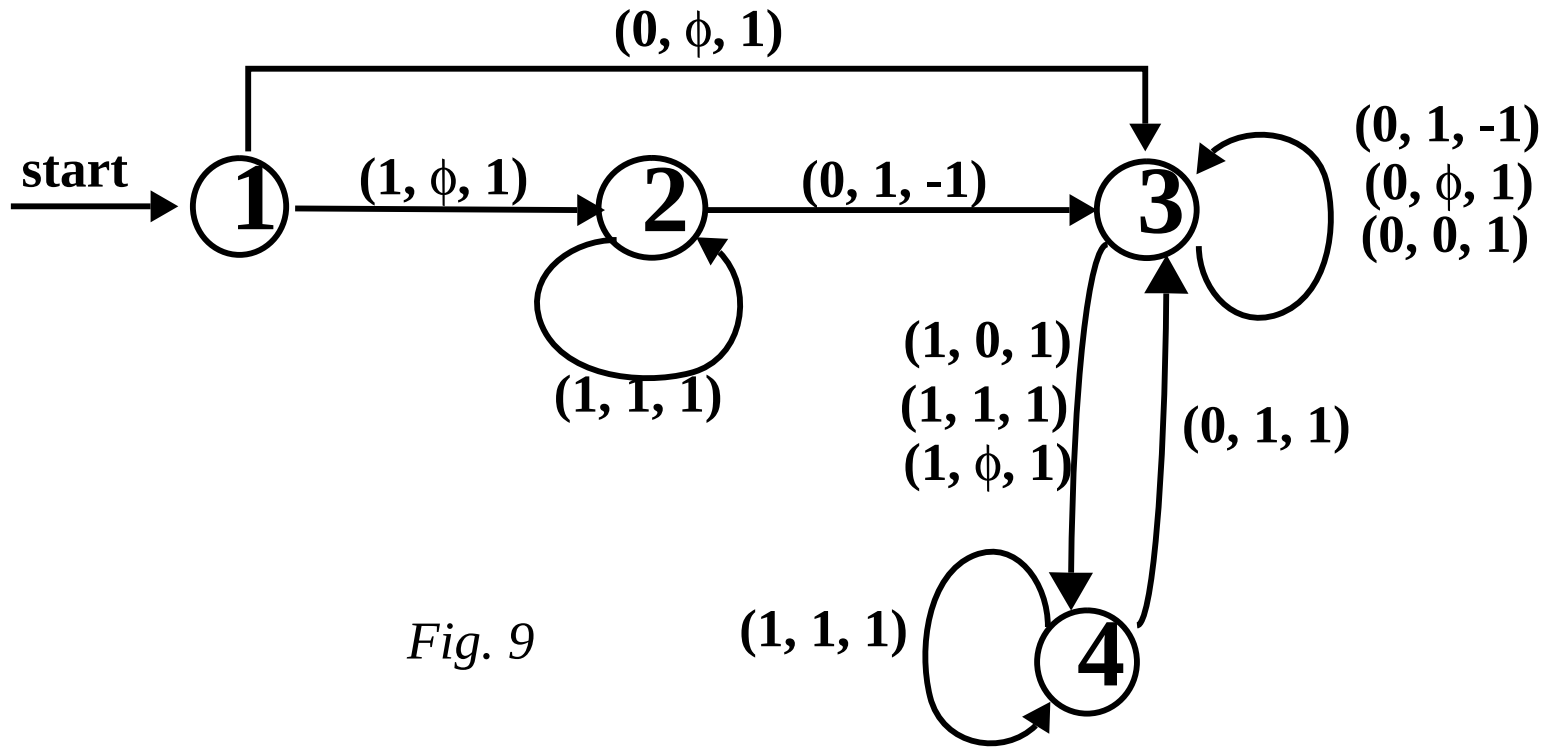

*Fig. 9*

Fig.9 The reduced PDA transition diagram of the $\mathbf{1^n0^n}$ grammar. This diagram is obtained by grouping together the equivalent states in Fig.8 and assigning one representation to each state group, where the states 1, 2, 3 and 4 represent respectively the quantized states (1,0,0,0,0), (1,0,0,0,1), the combination {(1,1,1,1,1), (1,0,1,1,1)} and the combination {(0,0,0,0,1), (0,0,0,1,1)}.

The second problem is the limitation of neural network structures. [Das92] shows that second and third order neural network structures are not able to learn certain grammars without "hints." Moreover, our simulations show that even with hints using second and third order networks, the palindrome grammar cannot be learned. The limitation of the neural network structure for learning the palindrome is now discussed. For example, the Palindrome grammar requires the action rules (a, a, 1) before seeing 'c' and (a, a, -1) after seeing 'c'. For these two rules, the input and the stack reading are the same but the action is different: one is push and the other is pop. So, according to the third order dynamics, the stack actions could be written $A = f(\boldsymbol{W} \cdot \boldsymbol{S} + \vartheta)$ where the summation over input symbols and stack readings for these two cases have already been performed and W is the result of the "equivalent weights". The problem becomes one of learning the weights $\boldsymbol{W}$ and $\vartheta$ such that $A$=1 for one set of states $\{\boldsymbol{S}_1\}$ (before seeing 'c') and $A$=-1 for another set of states $\{\boldsymbol{S}_2\}$ (after seeing 'c'). Clearly, two arbitrary sets of state vectors may not be linearly separable unless they all have a unary representation (or mutually orthogonal in general). (This is the assumption for justifying the usefulness of third order networks.) However, during learning the numerical neural states most likely to occur are neither unary nor mutually orthogonal. To overcome this problem we introduced the idea of a "full order" linear net for stack action mapping.

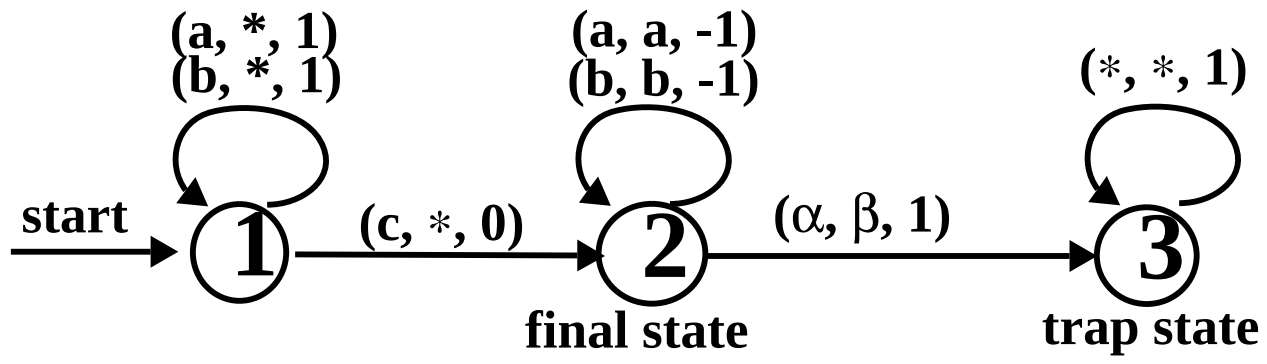

*Fig. 10*

Fig.10 The simplest PDA transition diagram for palindrome grammar, where α and β represent any combinations of input symbols and stack readings other than (a, a) and (b, b).

<u>(1). Full Third-order Network Structure.</u>

The third order connection weights for state dynamics as in Eq.(5) are used, and the stack action is governed by a linear "full order" mapping. The parameters are: (i) number of state neurons $N_S$=4, (Equivalent to the number of binary states = 16); (ii) Number of input symbol $N_I$=3, number of stack reading symbols $N_R$=4. Three input neurons for symbols 'a', 'b' and 'c' (no end symbol) and an additional neuron is introduced to represent the empty stack. This is necessary to supervise the learning to avoid the "empty stack" situation. (iii) One action neuron, $N_A$=1. In this case, the state transition weights as in Eq.(5) are a four-dimensional matrix $W^s[4][4][4][3]$ and the stack action weights are a three-dimensional matrix $W^a[16][4][3]$. The dynamics of the neural controller are

$$S_i^{t+1} = g\left(\sum_{j=1}^{N_S}\sum_{k=1}^{N_R}\sum_{l=1}^{N_I} W^s_{ijkl}\,(S_j^t R_k^t I_l^t) + \theta_i^s\right)$$

$$A^{t+1} = \sum_{J=1}^{2^{N_S}}\sum_{k=1}^{N_R}\sum_{l=1}^{N_I} W^a_{Jkl}\,(P_J^t R_k^t I_l^t) \quad , \qquad (23)$$

where the nonlinear function $f(x)$ in Eq.(8) has been replaced by a linear function $f(x) = x$ and the extended state vector $P_J$ is defined as

$$P_J^t = \prod_{m=1}^{N_S} \left(\delta_m S_m^t + (1-\delta_m)\,(1-S_m^t)\right). \qquad (24)$$

In Eq.(24), the symbol $\delta_m$ inside the product represents the binary values of 0 and 1, which are determined by the $m_{th}$ bit of the binary number (J-1). For example, if J-1 = 10, its binary form is 1010, which sets $\delta_m$: $\delta_1$=1, $\delta_2$=0, $\delta_3$=1 and $\delta_4$=0. The summation of all components of the extended state $P_J$ is equal to one, i.e.

$$\sum_{J=1}^{2^{N_S}} P_J^t = 1, \tag{25}$$

where $P_J$ can be interpreted as the probability for a NNPDA to be in each of the $2^{Ns}$ binary states. To guarantee that the action output be in the range: $-1 \le A^t \le 1$, the stack action weights are truncated to the range $-1 \le W^a \le 1$.

It can be seen that Eq.(25) plus the truncation of $W^a$ to [-1, 1] will automatically guarantee the action output in Eq.(23) to be within the range $-1 \le A^{t+1} \le 1$. Later, upon performing the post-learning quantization of $W^a$ to three levels: -1, 0 and 1, each of the action weights $W^a$ will represent an action rule, which were used in Figs.7- 10. For example, $W^a[3][2][1] = -1$ means that, starting from the third binary state, e.g. (0,0,1,0), if the input symbol is the first one, e.g. 0, and the stack reading is the second one, e.g. 1, the stack action will be a pop, i.e. a rule (0, 1, -1) marked besides the transition arrow from state (0,0,1,0) to the other state.

(2) Learning Criterion.

Some modifications have been made to the learning objective function previously discussed in Section 3.4. Both state and stack length are used to discriminate the legal and illegal strings. But, instead of using the usual desired final state and non-desired final state, we introduce the “trap state” and “non-trap state” to discriminate the “potentially legal string” and “definitely illegal string” [Das93]. Input strings “abbbacbab”, “abbbacbbababaaab”, ... , can now be classified before seeing the end of the string. This is because whenever symbol ‘b’ occurs after ‘c’, an ‘a’ in front of a ‘c’ is not matched and string is illegal irrespective of the remaining symbols. In that case, we force the NNPDA to go to the “trap state” and stop further learning. This requires prior knowledge about the underlying language in order to successfully supervise training. Here, we assigned the last state neuron to be 0 for the “trap state” and 1 for the “non-trap state”. For input strings not trapped into the “trap state,” training is as usual. The weight updates become

$$\Delta W = \eta \left[ (\boldsymbol{S}^* - \boldsymbol{S}^t) \frac{\partial \boldsymbol{S}^t}{\partial W} + (L^* - L^t) \frac{\partial L^t}{\partial W} \right] , \tag{26}$$

where $\boldsymbol{S}^*$ and $L^*$ are the target values of state and stack length. The target state is determined by the “trap state” or “non-trap state”, and the target stack length is zero for a legal string. Since the target stack length for an illegal string is not known, a small driving force is used empirically to slightly increase the stack length for all illegal strings ending at a “non-trap state”, i.e., $L^*-L^t = 0.1$ if $L^t \ge 0.9$ and $L^* = 1$ if $L^t < 0.9$. This error supervision is based on the following. Although the exact length of an illegal string is not known, it must be greater than or equal to one if the string ends up at a “non-trap state”. For illegal strings ending at a “trap state”, the stack length is unaffected.

(3)Training Set.

Two training sets are used. The first training set includes all 39 strings up to length three. The second contains 363 strings up to length five. Since the number of legal strings is much smaller than the illegal strings, the training set is balanced by adding all four legal strings up to length five to the first training set and all eight legal strings up to length seven to the second training set. In each training set the legal and illegal strings are put in two separate groups. During training, we present a legal string between every five illegal strings and make the learning rate for legal strings five times larger than that of illegal strings. Each training set was trained for 200 epochs.

(4)Training Algorithm.

The RTRL learning algorithm is generalized to the dynamics of Eqs.(23) to (26) which can be derived from the “chain rule” and forward propagating the error rate. Details are listed in Appendix B.

(5) Simulations of Training.

The first training set described above was used to train the NNPDA for 200 epochs. Then, the second training set was used for another 200 epochs. The “averaged classification error” for each training set was monitored during training. After a total 400 epochs of training, it converged to ~ 0.06. At the end of each string the error is determined by

$$E = (S^* - S^T_{Ns})^2 + (L^* - L^T)^2 \,. \tag{27}$$

The values of S* and $L^*$ are specified as before. The only difference from before is that for illegal strings the error ($L^*$ - $L^T$) is set to zero if $L^T$ is already greater than one.

The trained NNPDA is tested on new input strings. In testing the "trap state" monitor is not used to stop any sequence. The classifications criterion is: LEGAL if *both* $S^T_{Ns}$ >0.5 *and* $L^T \leq 0.5$; ILLEGAL otherwise. The 29,523 test strings include all possible strings constructed with symbol 'a', 'b' and 'c' up to length nine. [The following results are given for 5 significant figures, though 64bit floating point double precision was used.] The test result shows only four errors: three legal strings "ababcbaba" ($S^T_{Ns}$=0.9898, $L^T$=1.0776), "abbacabba" ($S^T_{Ns}$=0.9973, $L^T$=0.7301) and "bbbacabbb" ($S^T_{Ns}$=0.9994, $L^T$=0.5302) are classified as illegal because $L^T$>0.5 and one illegal string "abcbbbcbb" ($S^T_{Ns}$=0.9744, $L^T$=0.4543) is classified as legal because $S^T_{Ns}$>0.5 and $L^T$<0.5.

| Input string = "acabc", final stack length = 1.8805 > 0.5 -> classification Illegal. | | | | |
|---|---|---|---|---|
| input | state | action | stack segment lengths | stack symbols |
| a | (0.0079, 0.9952, 0.0160, 0.9580) | 1.0000 | (1.0000) | ( a ) |
| c | (0.0010, 0.0162, 0.9994, 0.9599) | 0.1323 | (1.0000, 0.1323) | ( a, c ) |
| a | (0.0026, 0.9982, 0.9971, 0.9995) | - 0.9869 | (0.1454) | ( a ) |
| b | (0.2055, 0.9749, 0.6775, 0.0003) | 0.7667 | (0.1454, 0.7667) | ( a , b ) |
| c | (0.0030, 0.9977, 0.4301, 0.9684) | 0.9684 | (0.1454, 0.7667, 0.9584) | ( a, b, c ) |

Table 1a.

| Input string = "bacab", final state = 0.9993 > 0.5, final stack length = 0.0318 < 0.5 -> classification Legal. | | | | |
|---|---|---|---|---|
| input | state | action | stack segment lengths | stack symbols |
| b | (0.9183, 0.0831, 0.9777, 0.9708) | 1.0000 | (1.0000) | ( b ) |
| a | (0.9934, 0.9875, 0.1103, 0.9999) | 0.9540 | (1.0000, 0.9540) | ( b, a ) |
| c | (0.0030, 0.1921, 0.9995, 0.9990) | 0.0625 | (1.0000, 0.9540, 0.0625) | ( b, a, c ) |
| a | (0.0021, 0.9989, 0.9961, 0.9998) | - 0.9989 | (1.0000, 0.0176) | ( b, a ) |
| b | (0.0031, 0.9089, 0.9994, 0.9993) | - 0.9858 | (0.0318) | ( b ) |

Table 1b.

| Input string = "bacba", final state = 0.0054 < 0.5, final stack length = 3.6539 > 0.5 -> classification Illegal. | | | | |
|---|---|---|---|---|
| input | Internal state | action | stack segment lengths | stack symbols |
| b | (0.9183, 0.0831, 0.9777, 0.9708) | 1.0000 | (1.0000) | ( b ) |
| a | (0.9934, 0.9875, 0.1103, 0.9999) | 0.9540 | (1.0000, 0.9540) | ( b, a ) |
| c | (0.0030, 0.1921, 0.9995, 0.9990) | 0.0625 | (1.0000, 0.9540, 0.0625) | ( b, a, c ) |
| b | (0.2890, 0.9472, 0.9021, 0.0260) | 0.6850 | (1.0000, 0.9540, 0.0625, 0.6850) | ( b, a, c, b ) |
| a | (0.0190, 0.99602, 0.4490, 0.0054) | 0.9524 | (1.0000, 0.9540, 0.0625, 0.6850, 0.9524) | ( b, a, c, b, a ) |

Table 1c.

*Table 1. A demonstration of the step by step working process of the trained NNPDA. The three example strings are "acaba", "abcba" and "abcab". The state of the NNPDA at each time step is displayed in each row using the data listed in the five columns. For all the cases, the initial neural state is (1, 0, 0, 0) and the initial stack reading is "empty stack". The first column is the input symbol, the second is the output of internal neural state $S^t$ represented as a four-dimensional vector, the third one is the action neuron output $A^t$, and the fourth and fifth are the stack status at each time step. The actual accuracy of the calculation was 64bit double-precision, but only 5 significant figures are shown.*

To illustrate the inner workings of the NNPDA for classification after training, consider the examples in Table 1, strings "acabc", "bacab" and "bacba". The processing status at each time step is displayed using the data listed in the five columns. For all the cases, the initial neural state is (1, 0, 0, 0) and the initial stack reading is "empty stack". At each time step the first, second, third, fourth and fifth columns are the input symbol, the four-dimensional neural state $\boldsymbol{S}^t$, the action neuron output $A^t$, and the stack segment length and symbol, respectively. For example, the combination of (1.0000, 0.1323) in the fourth column and (a, c) in the fifth column represent a stack configuration: symbols 'a' at the bottom with length = 1.0000 and 'c' at the top with length = 0.1323.

See the first example in Table 1a. The whole string is an illegal pattern "acabc", but the first three symbol consists of a legal string "aca". When "aca" is fed in, the trained NNPDA first pushes 'a' with length 1.0000 into the stack, then pushes again the second input symbol 'c' with length 0.1323 and finally pops the stack with total length 0.9869. In the stack remains a symbol 'a' with final length $L^T$= 0.1454(<0.5). The internal state varies and reaches a final state such that neuron $S^T_{Ns}$=0.9995(>0.5). Therefore, the string "aca" is classified as legal ($S^T_{Ns}$>0.5 and $L^T$<0.5). Notice that all three states are a "non-trap state" (because $S^T_{Ns}$ > 0.95 for all cases). But, when an additional symbol 'b' is read, the state changed to a "trap state" indicating that "acac" is an illegal string. During the training we ignored the rest of the sequence and concluded that no matter what the next symbol, the entire string would be illegal. But, in the test sequence, the "trap state" monitor is not used and classification of any strings will be decided at the end of each string. After feeding in another symbol 'c', the state becomes a "non-trap state" (not a desired state). Fortunately, the stack actions in the last two step are pushes and the final stack length $L^T$=1.8805>0.5, classifying the entire string as illegal.

In Table 1b, the trained NNPDA deals with a legal string "bacab" nearly perfectly. The controller first pushes 'b' and 'a' onto the stack and then moves to a special state (0.0030, 0.1921, 0.9995, 0.9990) after seeing 'c' (but does not push much of 'c' into the stack since 0.0625 is a tolerable error). It pops 'a' and 'b' out of the stack when the input symbol matches the stack readings. Concurrently, the state remains in the "non-trap state" as desired.

The Table 1c shows what happens if we reverse the order of last two symbols 'a' and 'b' in the last example. Again, the trained NNPDA behaves nearly perfectly. When the fourth symbol 'b' is fed in, the stack reading is almost a complete 'a' (a combination of 'c' with length 0.0625 and 'a' with length 0.9375). Since the input 'b' does not match the stack reading 'a', the NNPDA enters a "trap state" and the string "bacb" is classified as illegal. Furthermore, if another symbol 'a' is seen, the NNPDA moves to another "trap state". So, "bacba" is still illegal. Concurrently, the stack actions generated from the "trap state" are all pushes. These increase the stack length so that the classification is "far" from legal.

Although the classifications for these three examples are all correct, in the sense of a correct discrete PDA, there are still some numerical errors. These numerical errors will accumulate over time and possibly misclassify an input string that is too long. One of the four incorrect classifications in our test result, the string "ababcbaba", is illustrated in Table 2, where the general behavior of the learned NNPDA is the same as that of a discrete PDA. But, due to the accumulation of numerical errors, at t =7 when the input symbol is 'a', the NNPDA reads not a complete 'a' in the stack. Instead, it reads with depth unity an 'a' with length 0.6467 and a 'b' with length 0.3533. Therefore, the action output is not a full "pop" but a "pop" with length 0.2757. Thus, accumulated final stack length is 1.0776 > 0.5 and the string is classified as illegal.

(6). Quantization of the Trained NNPDA.

The state neuron activities are quantized to two levels. The stack action weights $\boldsymbol{W}^a$ are quantized to three levels:

$$W^a = \begin{cases} 0, & \text{if} \quad (|W^a| \le 0.5) \\ -1, & \text{if} \quad (W^a < -0.5) \\ 1, & \text{if} \quad (W^a > 0.5) \end{cases} . \tag{28}$$

After quantization, we test the NNPDA with all possible strings up to length fifteen. The classification rule is as follows. The "trap state" monitor is used to monitor the last state neuron $S^t_{Ns}$. Whenever $S^t_{Ns}$ becomes zero, we stop the sequence and classify it as an illegal string; otherwise, we proceed to the end of the input sequence. At the end, if $L^T$=0, the input is classified as legal; otherwise it is illegal. The test result is that all the 21,523,359 strings are classified correctly. But, this does not mean that the quantized NNPDA represents the Palindrome grammar. We have to extract the correct discrete PDA and verify that it recognizes the Palindrome grammar.

(7). Extraction of the Discrete PDA.

| Input string = “ababcbaba”, final stack length = 1.0776 > 0.5 -> classification Illegal. | | | | |
|---|---|---|---|---|
| input | Internal state | action | stack segment lengths | stack symbols |
| a | (0.0077, 0.9952, 0.0160, 0.9580) | 1.0000 | (1.0000) | ( a ) |
| b | (0.9855, 0.9364, 0.9868, 0.9784) | 0.9716 | (1.0000, 0.9716) | ( a, b ) |
| a | (0.9627, 0.9961, 0.1055, 0.9811) | 0.9936 | (1.0000, 0.9716, 0.9936) | ( a, b, a ) |
| b | (0.9987, 0.8105, 0.9719, 0.9995) | 0.9932 | (1.0000, 0.9716, 0.9936, 0.9932) | ( a, b, a, b ) |
| c | (0.0002, 1.0000, 0.0239, 0.9992) | 0.0810 | (1.0000, 0.9716, 0.9936, 0.9932, 0.0810) | ( a, b, a, b, c ) |
| b | (0.0053, 0.9977, 0.9881, 0.9996) | - 0.9981 | (1.0000, 0.9716, 0.9936, 0.0761) | ( a, b, a, b ) |
| a | (0.0016, 0.9996, 0.9209, 0.9993) | - 0.8207 | (1.0000, 0.9716, 0.2491) | ( a, b, a ) |
| b | (0.0246, 0.9377, 0.9986, 0.9937) | - 0.8674 | (1.0000, 0.3533) | ( a, b ) |
| a | (0.0128, 0.9994, 0.7910, 0.9898) | - 0.2757 | (1.0000, 0.0776) | ( a, b ) |

Table 2

*Table 2. The step by step operations of a numerically trained NNPDA for the example string “ababcbaba”. The general behavior of the analog NNPDA is correct. But, due to the cumulated numerical round-off error, the action output deviates gradually from the discrete pop so that the final classification is wrong.*

Using the quantized NNPDA with the initial state (1, 0, 0, 0), we check all possible paths of the quantized NNPDA by reading input symbols as described in Section 4.1. The transition diagram of these paths is drawn in Fig.11. Every path was terminated whenever a “trap state” occurred. Each bracketed action rule in the form of (input, reading, action) is marked besides the transition arrows. This diagram looks more complicated than might be expected. Though it did not turn out to be the simple diagram of Fig.10, the neural net generates some rather novel transitions.

First we find all equivalent states. All “trap states” are equivalent. Also, the two states (1,0,1,1) and (0,0,1,1) make equivalent transitions and actions. After grouping these equivalent states, seven states are finally selected and labelled as in Fig.12. The first six states are “non-trap states” and the seventh is the “trap state”. Let us see how this PDA shown in Fig.12 could recognize the Palindrome grammar. The start state is state (1) and the start reading is an “empty stack” represented by ‘$\phi$’. If the first input symbol is ‘c’, it will move to state(3), and then either stop there with an empty stack (the string “c” is legal) or goes to the “trap state”—state(7) if more symbols are read (i.e. an illegal string). When an input string starts with ‘a’ or ‘b’, the neural net controller pushes the read symbol onto the stack and moves to either state(2) (for input ‘b’) or state(3) (for input ‘a’). Then, before seeing a symbol ‘c’, it will push all the read symbols onto the stack and, concurrently, move among a symmetric structure of the four states (2), (3), (4) and (5). These four states are manipulated in a very complicated manner. Whenever a symbol ‘b’ is read, it is pushed onto the stack and the PDA moves to either state(2) or state(5). Whenever a symbol ‘a’ is read, it is pushed onto the stack and the PDA moves to either state(3) or state(4). If a symbol ‘c’ is read, it will transit to either state(2) (if the last symbol is ‘a’) or state(3) (if the last symbol is ‘b’). Then, the controller will examine whether the next input symbol matches the top symbol on the stack. If every read symbol matches the stack reading, the PDA will pop and move to state(6) and stay there until the stack is emptied. If any input symbol does not match the top stack symbol. the PDA will go to the “trap state”— state(7) and the string is classified as illegal.

As noted in Fig.12, the self-loop for state(7) indicates that there is no escape from a “trap state.” This is assumed because of our pre-knowledge about the “trap state”. However, the discrete NNPDA-generated “trap states” may not form closed loops. We have checked all the possible transitions from the “trap states” and find that there do exist “leaks”. For example, the illegal string “bbcbabacabab” is found to end at a “non-trap state” (1,1,1,1) and the string “aaacabbcb” ends at (1,0,1,1). Thus, it is good idea to use the “trap state” monitor in recognition as well as in training.

# V. CONCLUSION

A recurrent neural network pushdown automata (NNPDA) was devised and used to learn simple but illustrative deterministic context-free grammars (CFGs). The NNPDA itself is a hybrid model consisting of a recurrent neural network state automaton controller and an external *continuous* stack memory connected through a common error function. This is to be contrasted to connectionist models that construct stacks (and their associated state structure) from internal hidden layers or from the dynamic range of the nonlinearity of the neural network. To train the NNPDA an enhanced forward-propagating real time recurrent learning algorithm (RTRL) was derived and used to learn CFGs from positive and negative string examples. However, the NNPDA model is quite general and can be trained using other gradient descent approaches such as a modified back-propagation through time algorithm. What should be noted is that during training the NNPDA *simultaneously* learns to construct its internal state controller and to figure out how to control with the proper actions (push, pop and no-operation) the use of the external stack memory.

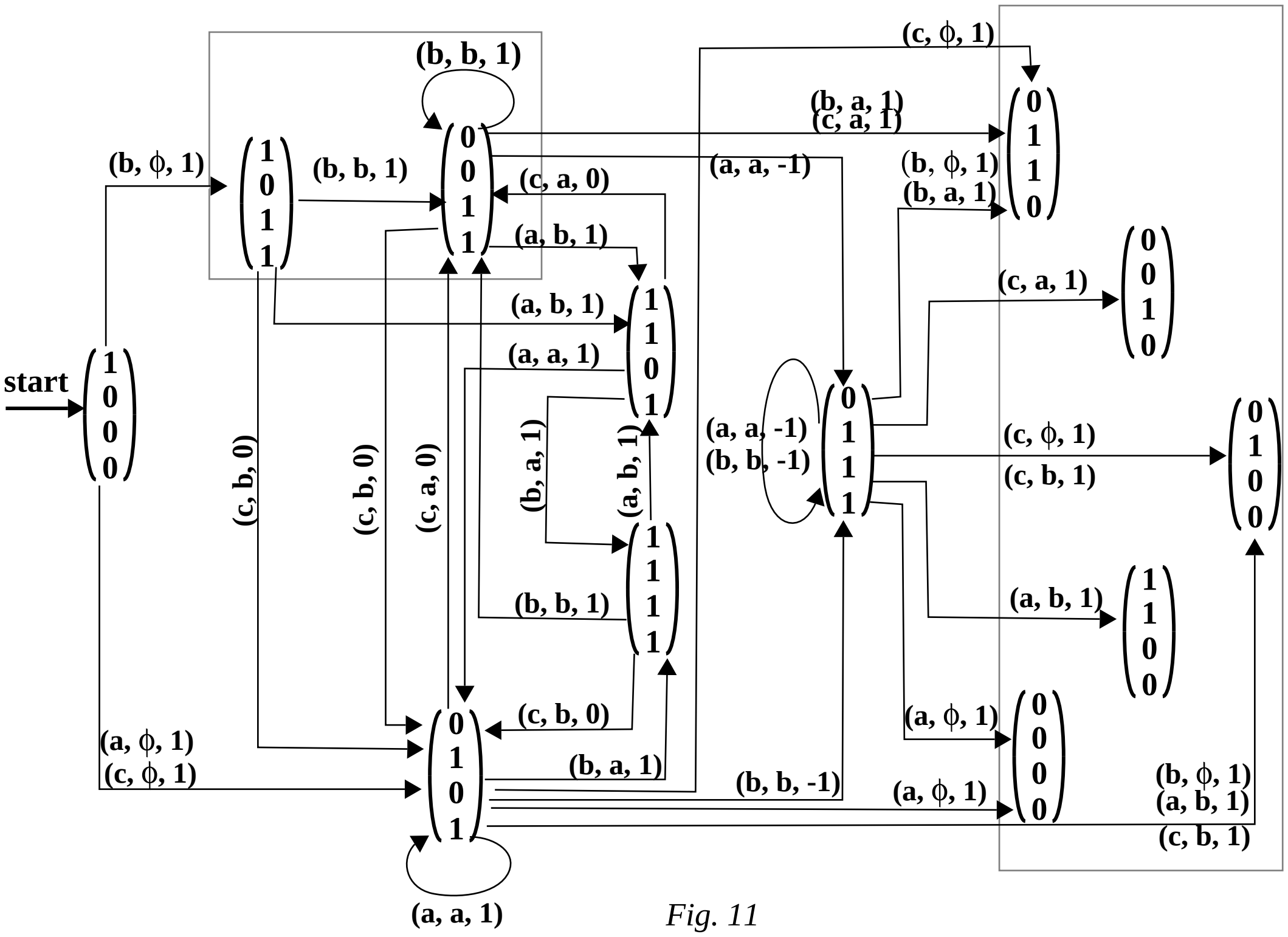

*Fig. 11*

Fig.11 The extracted discrete PDA obtained from the trained NNPDA by quantization of the neural activities of the continuous NNPDA. Using the quantized NNPDA, start with the initial state (1, 0, 0, 0) and cover all possible paths by feeding in various strings whenever needed. Here, all paths are terminated whenever a "trap state" occurs. Each bracketed action rule in the form of (input, reading, action) is marked by the transition arrows.

The external continuous stack memory is constructed of two arrays; one for symbols and one for real values associate with those symbols. The input symbol alphabet is also the stack alphabet (this somewhat restricts the class of learnable CFGs). A gradient-descent training algorithm is derived for the continuous stack. One interpretation of the continuous stack memory is that the real values associated with the symbols stored on the stack reflect an uncertainty in the content of stack reading of the NNPDA. This allows more than one symbol to be read from the top of stack and each with different probabilities.

For all languages of the learned grammars (the balanced parenthesis, $1^n0^n$ and palindrome grammars), the size of the positive and negative string training set was less than 512. The number of epochs required for successful training was approximately 100 and usually less than 1000. The trained NNPDA exhibited very good generalization capabilities and were able to correctly classify large sets (usually millions) of unseen strings. Its performance appears to be much better than other connectionst stack models used to learn simple context-free grammars.

We devised an algorithm for extracting a discrete pushdown automaton (PDA) from the trained NNPDA. For all the grammars used in training, correct PDAs were extracted (For all languages the strings were generated by "known" PDAs). The advantage of this quantization process is that the extracted PDA was often able to outperform the trained

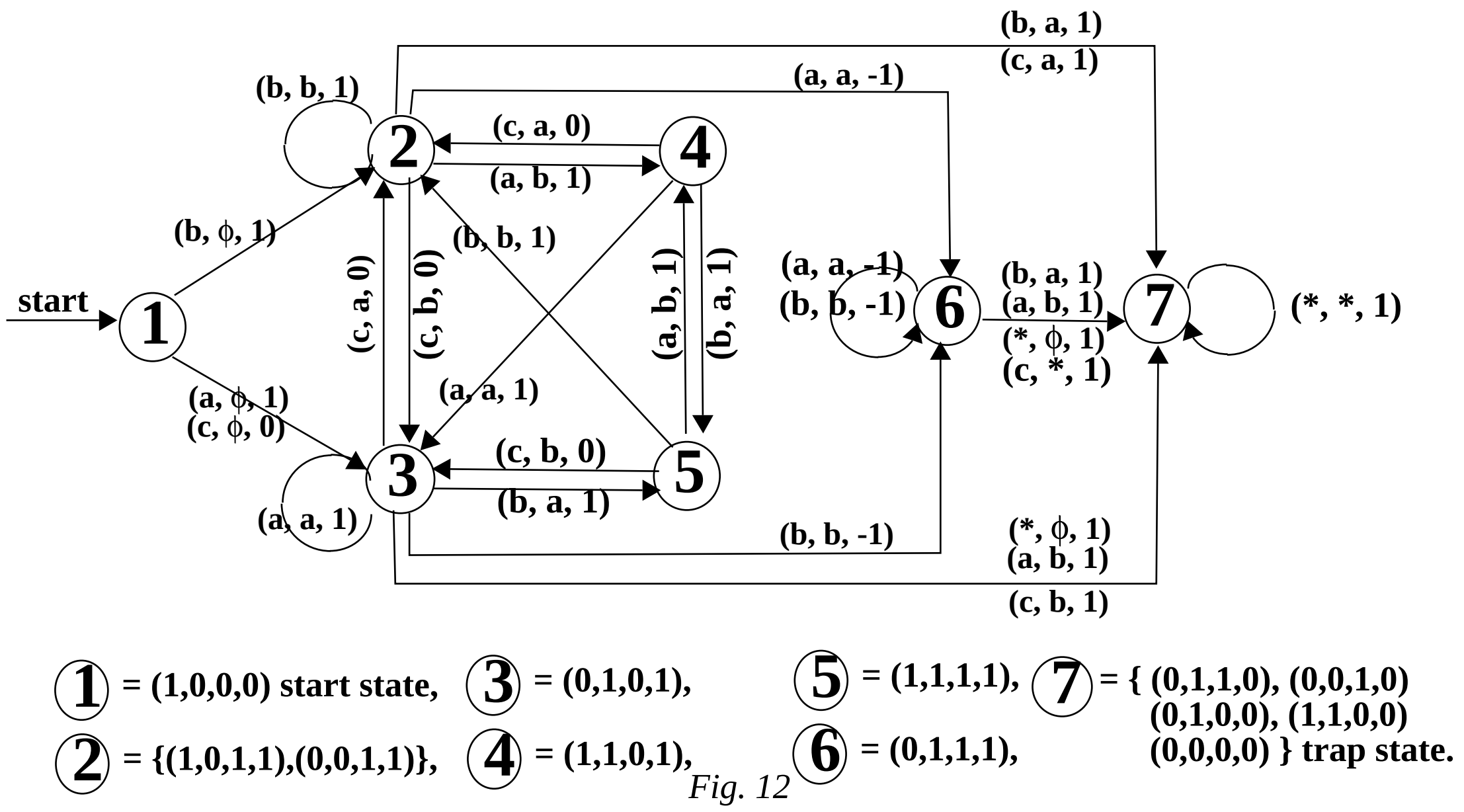

Fig.12 The equivalent reduced PDA that recognizes the palindrome grammar. It is obtained by grouping the equivalent states of the PDA in Fig.11 into seven representative states and completing their transitions. The correspondence between the original 12 states and the reduced 7 states is listed in the seven equalities below the transition diagram.

NNPDA in correctly classifying any unseen strings (similar results were shown for FSA extracted from trained NNFSA [Giles92a, Omlin92]). However, the extracted PDAs could be quite complex and not necessarily the simple PDA expected.

There are many open issues. We only demonstrated the principle of simultaneously training a recurrent neural network coupled to an external stack memory. It is not evident that this method will scale or this is an efficient way to learn context free grammars. There needs to be further work on the required accuracy of the analog stack. The additional knowledge required to learn the palindrome grammar shows that the intelligent use of topology, such as order of connection weights, and *a priori* knowledge, such as supervising the control of the stack, significantly effects successful training and testing. Because of the number of variables, the training results were illustrative not exhaustive or complete. What was interesting is that such good results were obtained!

Finally, there is nothing that restricts this model to symbol learning. Real numbers could have just as easily been used as inputs. We speculate that this model could also be used in learning more complex hidden state processes for real-valued problems.

# Acknowledgments

The University of Maryland authors gratefully acknowledge partial support by AFOSR and ARPA.

# Appendix A

# The derivation of $\partial R^t / \partial W$

In this appendix we derive $\partial R^t_{k'} / \partial W_{ijk}$ for the case where there is only one action neuron $N_A$=1. The generalization to the case with more action neurons is straightforward.

The stack reading at time t is in general a function of the entire stack history

$$R^t = F(A^1, A^2, \ldots, A^t, I^1, I^2, \ldots, I^t) \quad , \tag{A-1}$$

where $A^\tau \in [-1, 1]$, $1 \le \tau \le t$, is the continuous action value which operates on the stack. The input symbol $I^\tau$, $1 \le \tau \le t$, at time $\tau$ is read from the input sequence. As previously defined, an action to be performed on the stack is either a *push, pop* or *no-operation (no-op)* depending on the sign and magnitude of $A^\tau$. The amount of the stack to be *push*ed or *popped* is equal to the absolute value of $A^\tau$, which also determines what amount of that the current input symbol $I^\tau$ is read into the stack.

To complete the forward-propagation of the sensitivity matrices $\partial \boldsymbol{S}^t / \partial \boldsymbol{W}$ and $\partial \boldsymbol{A}^t / \partial \boldsymbol{W}$ as in Eq.(19), the derivative $\partial \boldsymbol{R}^t / \partial \boldsymbol{W}$ has to be known. If a recursive relation for $\partial \boldsymbol{R}^t / \partial \boldsymbol{W}$ exists, *i.e*.

$$\frac{\partial \boldsymbol{R}^{t+1}}{\partial \boldsymbol{W}} = \boldsymbol{M}\left(\frac{\partial \boldsymbol{R}^t}{\partial \boldsymbol{W}}, \frac{\partial \boldsymbol{S}^t}{\partial \boldsymbol{W}}, \frac{\partial \boldsymbol{A}^t}{\partial \boldsymbol{W}}, \boldsymbol{I}^t\right), \tag{A-2}$$

where $\boldsymbol{M}$ is an unknown vector function, the recursive evaluation of $\partial \boldsymbol{S}^t / \partial \boldsymbol{W}$ and $\partial \boldsymbol{A}^t / \partial \boldsymbol{W}$ is straightforward. However, a rigorous recursion equation of Eq.(A-2) does not exist. The reason is as follows.

The stack operation and stack reading $\boldsymbol{R}^t$ defined in Section III does not include any derivative of $\boldsymbol{R}^t$ with respect to $\boldsymbol{W}$. Therefore, Eq.(A-2) implies the following relation

$$\boldsymbol{R}^{t+1} = \boldsymbol{H}(\boldsymbol{R}^t, \boldsymbol{S}^t, \boldsymbol{A}^t, \boldsymbol{I}^t) \quad , \tag{A-3}$$

where $\boldsymbol{H}$ is another vector function. But, in general, relation (A-3) should not hold for a PDA. The reason is that the current stack reading $R^t$ depends on the whole history of the stack, not on the history a few time steps in the past. If, we assume that relation (A-3) is true, then the read operation can couple with the dynamics of the neural network controller, as in the two equations in Eq.(3). This yields

$$\boldsymbol{Z}^{t+1} = \boldsymbol{K}(\boldsymbol{Z}^t, \boldsymbol{I}^t) \quad , \tag{A-4}$$

where the vector $\boldsymbol{Z}$ represents the concatenation of the three vectors $\boldsymbol{R}$, $\boldsymbol{S}$, and $\boldsymbol{A}$, or $\boldsymbol{Z} \equiv (\boldsymbol{S} \oplus \boldsymbol{A} \oplus \boldsymbol{R})$, and $\boldsymbol{K}$ is the combination of the functions: $\boldsymbol{H}$ in (A-3), $\boldsymbol{G}$ and $\boldsymbol{F}$ in Eq.(3). Since in the discrete limit the vector $\boldsymbol{Z}$ is represented by finite description, the relation of Eq.(A-4) indicates that the whole system is a finite state automaton with extended internal states represented by $\boldsymbol{Z}$.

The fallacy of assuming that Eq.(A-3) is correct can also be seen from a simple example. Suppose that the input sequence contains 20 symbols and the stack is empty. The PDA is constrained to have two actions: from t = 1 to t = 10 only pushes and from t = 11 to t = 20 only pops. Then, after the nineteenth action (pop) there would be only one symbol left on the stack. The content of the stack reading $\boldsymbol{R}^{20}$ is the first symbol of the input string pushed onto the stack at time t = 1. This is a counter-example to (A-3), since $\boldsymbol{R}^{20}$ not only depends on the previous reading $\boldsymbol{R}^{19}$, previous action $\boldsymbol{A}^{19}$ and state $\boldsymbol{S}^{19}$ but also on $\boldsymbol{I}^1$ and $\boldsymbol{A}^1$, the stack history at time t= 1.

Generally speaking, the exact calculation of $\partial R^{t+1}_{k'} / \partial W_{ijk}$ will involve the storage of the entire history of the stack and actions on the stack, which demands a large memory size and increased computation. In order to simplify this problem, we derive an efficient approximation to $\partial R^{t+1}_{k'} / \partial W_{ijk}$ which can be used recursively in a manner that closely approximates the recursion set of Eq.(19). Since the input symbol $\boldsymbol{I}^t$ does not depend on the weight $\boldsymbol{W}$, Eq.(A-1) implies that

$$\frac{\partial \boldsymbol{R}^t}{\partial \boldsymbol{W}} = \sum_{\tau=1}^{t} \frac{\partial \boldsymbol{R}^t}{\partial \boldsymbol{A}^\tau} \cdot \frac{\partial \boldsymbol{A}^\tau}{\partial \boldsymbol{W}}, \tag{A-5}$$

where the summation over $\tau$ in general contains all time steps starting at t = 1. But not all of the history of $\boldsymbol{A}^\tau$ affects the current stack reading $\boldsymbol{R}^t$. Since $\boldsymbol{R}^t$ contains only the contents of depth 1 from the top of the stack, the number of terms in the summation (A-5) can be reduced by removing all of the actions $\{\boldsymbol{A}^\tau, 1 \le \tau \le t\}$ which do not contribute to the generation of $\boldsymbol{R}^t$.

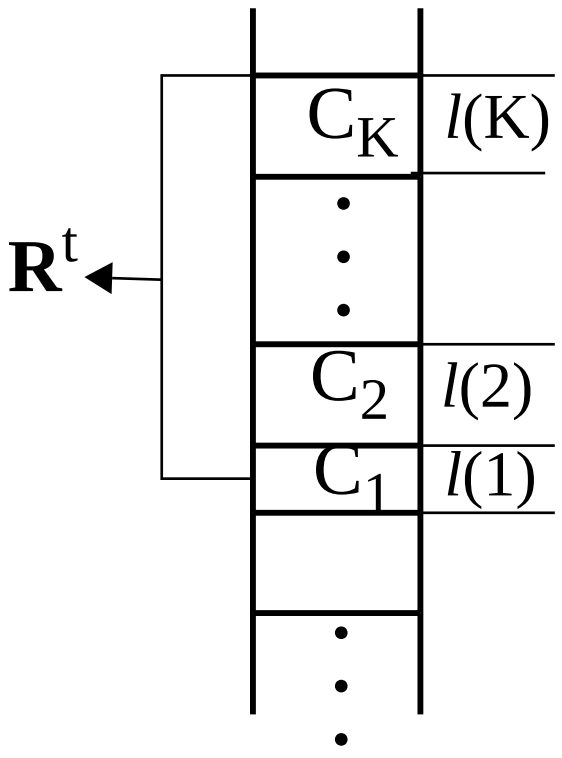

*Fig. A1*

Fig.A1 The reading $\mathbf{R}^t$ of the continuous stack at time t consists of K sections of continuous symbols, $C_i$, i = 1, 2, ... K, each of which contains only one symbol, and each pair of adjacent sections, $C_i$ and $C_{i+1}$, contains two different symbols. The length of each section $C_i$ is denoted by l(i) marked beside the stack.

Assume that $\boldsymbol{R}^t$ consists of $K$ sections of continuous symbols, as shown in Fig. A1, where each section contains only one symbol denoted by $C_i$ with length $l$(i), each generated at time $\tau_i$, i = 1, 2,... $K$. Each pair of adjacent sections, $C_i$ and $C_{i+1}$, contains different symbols. Note that each of the sections $\{C_i,\ i=1,\ 2,\ ...\ K\}$ may not be generated by only one action (*push*) at time $\tau_i$. It may be first generated partially at time $\tau_i$ and then be *popped* or *pushed* several times. Finally, (before time $\tau_{i+1}$ when the next symbol $C_{i+1}$ was generated) the symbol $C_i$ with length $l$(i) is left on the stack. Under these assumptions, the actions before $\tau_1$: $\{\boldsymbol{A}^\tau,\ 1 \le \tau < \tau_1\}$ do not contribute to the formation of $\boldsymbol{R}^t$, and therefore can be removed from the summation. The expression Eq.(A-5) can be written as

$$\frac{\partial \boldsymbol{R}^t}{\partial \boldsymbol{W}} = \sum_{i=1}^{K} \sum_{\tau_i \le \tau < \tau_{i+1}} \frac{\partial \boldsymbol{R}^t}{\partial A^\tau} \cdot \frac{\partial A^\tau}{\partial \boldsymbol{W}}\ , \qquad \text{(A-6)}$$

where the bold-faced $\boldsymbol{A}^\tau$ have been replaced by $A^\tau$ (without lose of generality, only one action neuron is used).

In order to calculate the derivatives in Eq.(A-6), assume that there is an infinitesimal perturbation of the weight matrix $\Delta\mathbf{W}$**,** which then produces infinitesimal perturbations of actions $\{\Delta A^\tau,\ 1 \le \tau < t\}$ for every time step calculated from the second equation in Eq.(3). These new actions $\{A^\tau + \Delta A^\tau,\ 1 \le \tau < t\}$ can be used to reconfigure the stack, which in turn creates the change in the stack reading $\Delta\boldsymbol{R}^t$. The partial derivative $\partial R^t_k / \partial A^\tau$ is defined as

$$\frac{\partial R^t_k}{\partial A^\tau} = \lim_{\Delta A^\tau \to 0} \frac{\Delta R^t_k}{\Delta A^\tau}\ , \qquad \text{(A-7)}$$

where the change in the stack reading $\Delta R^t_k$ is induced by only $\Delta A^\tau$, while all other $A^{\tau'}$, $\tau$ ś $\neq \tau$, are fixed.

Since the stack reading $\boldsymbol{R}^t$ consists of $K$ sections of continuous symbols $\{C_i,\ i=1,\ 2,\ ...\ K\}$, the change $\Delta\boldsymbol{R}^t$ would also be computed from $\{\Delta C_i,\ i=1,\ 2,\ ...\ K\}$, the change in each of the sections. The major approximation made in this derivation is the following. We assume that for $\tau_i \le \tau < \tau_{i+1}$, an infinitesimal perturbation $\Delta A^\tau$ would only produce the change of the length, $\Delta l$(i), in the $i_{th}$ section $C_i$. In general, this is not true, because there exists a perturbation $\Delta A^\tau$ which not only changes the length of its symbol section but also changes the content of the section (i.e. brings in a part of the new symbol to this section). This can be seen from a counter example. Suppose that the section $C_i$ contains only the symbol ***a*** with length 0.5 and it is produced by a sequence of actions: (1) at $\tau=\tau_i$, $A^\tau$=0.1 (*push*) of symbol ***a***, (2) at $\tau=\tau_i+1$, $A^\tau$=0.2(*push*) of symbol ***b***, (3) at $\tau=\tau_i+2$, $A^\tau$= -0.2 (*pop*). (4) at $\tau=\tau_i+3$, $A^\tau$=0.4 (*push*) of symbol ***a***. Although the net result is to push a symbol ***a*** with length 0.5 onto the stack, during the sequence an equal amount of symbol ***b*** was pushed and popped onto and from the stack. In this case an infinitesimal perturbation $\Delta A^\tau>0$ when $\tau=\tau_i+1$ or $\tau_i+2$ would create an infinitesimal portion of symbol ***b*** with length equal to the absolute value of $\Delta A^\tau$, sandwiched between the two parts of symbol ***a***. We ignore this situation because of the following reasoning.

Assign an *occurrence probability* $\rho(A^\tau)$ to each action $A^\tau$ and replace the derivative $\partial R^t_k/\partial A^\tau$ by its probability weighted value:

$$\frac{\partial R_k^t}{\partial A^\tau} \to \frac{\partial R_k^t}{\partial A^\tau}\rho\,(A^\tau)\ . \tag{A-8}$$

If an action $A^\tau$ is free to take any values in the domain [-1, 1], assign it *occurrence probability* one. In Eq.(A-6) all the terms on the right hand side of the summation are supposed to have a value of *occurrence probability* equal to one. However, we argue that there do exist some actions with zero *occurrence probability* and that these can be removed from the summation in Eq.(A-6). A special group of such actions are the *pops* which pop symbols at their *boundaries*, i.e. the border lines inside the stack which separate two different symbols. For instance, in the above example the action $A^\tau= -0.2$ (*pop*) at $\tau=\tau_i+2$ belongs to this category. In general, for the stack example shown in Fig.A1, if the next time action is $A^{t+1}=-\mathrm{B(i)}$, $\mathrm{B(i)}\equiv l(K)+l(K-1)+...+l(\mathrm{i}+1)+l(\mathrm{i})$ for any $\mathrm{i}=1,2,3,...,K$, we would say that the action $A^{t+1}$ has a zero *occurrence probability*. In fact, if the action $A^{t+1}$ is uniformly distributed within [-1, 1], then the *occurrence probability* of $A^{t+1}$ can be measured by the possible range of values of $A^{t+1}$ divided by 2, the measure of whole region [-1, 1]. If $A^{t+1}$ is a *pop* which occurred around a boundary of one of the stack sections shown in Fig.A1, say $|A^{t+1}+\mathrm{B(i)}|\le\varepsilon$, then the measure of the *occurrence probability* of $A^{t+1}$ will be $\varepsilon/2$. When $\varepsilon\to 0$ (or when $A^{t+1}\to -\mathrm{B(i)}$, $\mathrm{i}=1,2,3,...,K$) this probability goes to zero.

With the above approximation we have two useful outcomes. First, in the summation on the right-hand side of Eq.(A-6), all terms within the first section whose action $A^\tau$ occurs between time $\tau_1$ and $\tau_2$, i.e. $\tau_1\le\tau<\tau_2$, can be removed and the equation becomes

$$\frac{\partial \boldsymbol{R}^t}{\partial \boldsymbol{W}} = \sum_{i=2}^{K}\ \sum_{\tau_i\le\tau<\tau_{i+1}} \frac{\partial \boldsymbol{R}^t}{\partial A^\tau}\cdot\frac{\partial A^\tau}{\partial \boldsymbol{W}}\,; \tag{A-9}$$

because for all actions $\{A^\tau, \tau_1\le\tau<\tau_2\}$, $\partial\boldsymbol{R}^t/\partial A^\tau$ are zero. The reason is that the content of the stack reading $\boldsymbol{R}^t$ is formed by reading the stack in the top-down manner with a fixed length 1 and is actually independent of the infinitesimal change of the length on the first section. As shown in Fig.A2, as long as the lower boundary of section $C_1$ does not exactly coincide with the lower boundary of $\boldsymbol{R}^t$, the content of $\boldsymbol{R}^t$ will not change. In the case where the lower boundary of section $C_1$ does coincide with the lower boundary of $R^t$, any negative change of the section $C_1$'s length ($\Delta l<0$) will introduce an infinitesimal change in $\boldsymbol{R}^t$. This case was excluded because it has zero *occurrence probability*.

From Figure A2, a method for determining $\partial\boldsymbol{R}^t/\partial A^\tau$ within each section $C_i$, i = 2,3,..., k, can be calculated as follows. According to the definition in Eq.(A-7), we need to calculate the ratio $\Delta R^t_k/\Delta A^\tau$ and take the limit $\Delta A^\tau\to 0$. Suppose $\tau_i\le\tau<\tau_{i+1}$. It is known that the perturbation $\Delta A^\tau$ only changes the length of the section $C_i$. But, the stack reading $\boldsymbol{R}^t$ will still have a fixed length (a distance of one) regardless of this change. Therefore, the contents of the stack reading would not only include $\Delta l(\mathrm{i})$, the change in the length of symbol $C_i$, but also $-\Delta l(1)$, the change of symbol $C_1$. See Fig.A3. This implies

$$\frac{\partial R_k^t}{\partial A^\tau} = \lim_{\Delta A^\tau\to 0}\frac{\Delta R_k^t}{\Delta A^\tau} = \lim_{\Delta A^\tau\to 0}\frac{1}{\Delta A^\tau}\left(\Delta l\,(i) - \Delta l\,(1)\right)_k \quad .$$

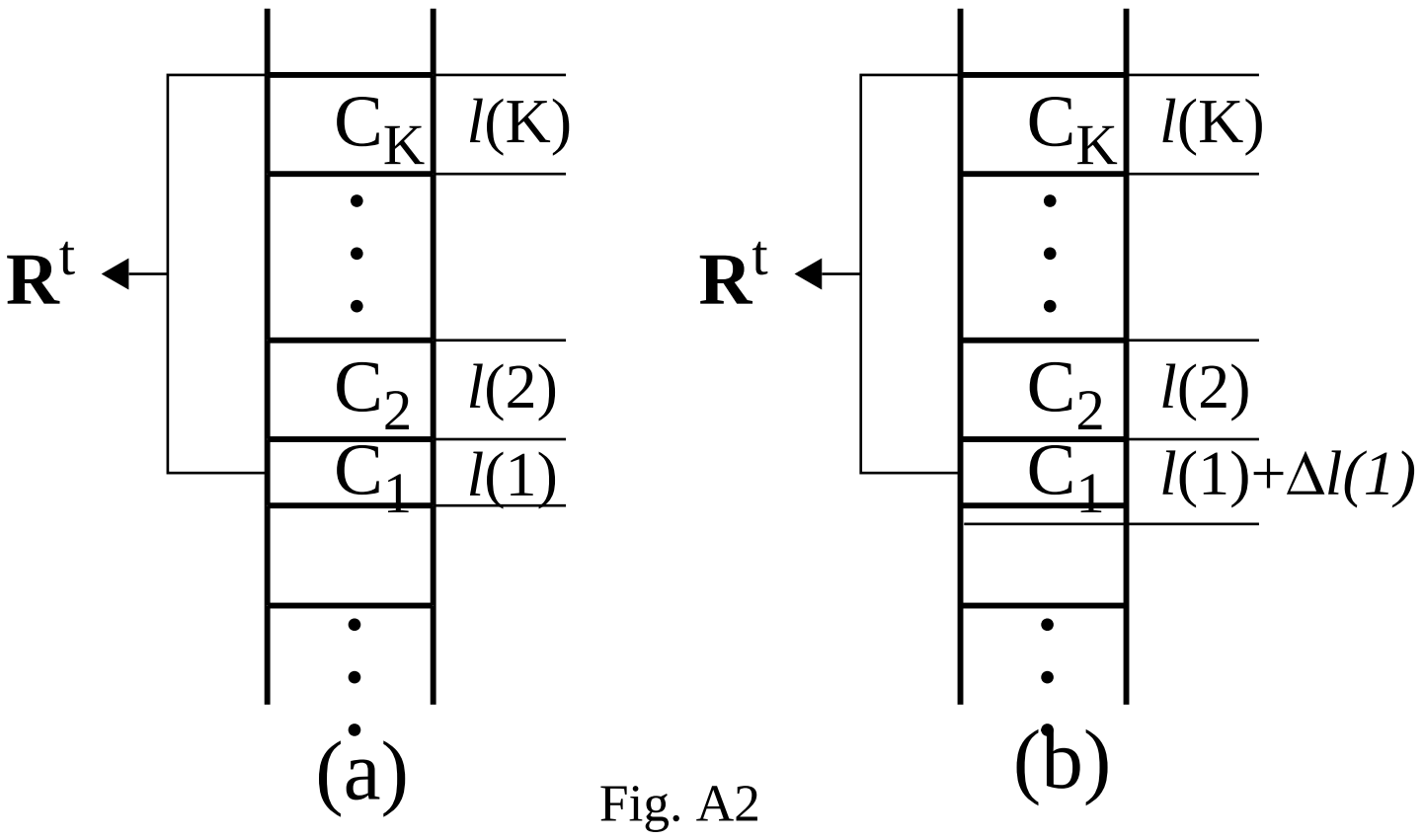

Fig. A2

Fig.A2 When the stack reading $\mathbf{R}^t$ has a fixed length = 1, its content is independent of $\Delta l(1)$, an infinitesimal change of the length in first section $C_1$, unless (i) the lower boundary of section $C_1$ coincides with the lower boundary of $\mathbf{R}^t$ and (ii) $\Delta l(1) < 0$.

(a) Stack reading $\mathbf{R}^t$ before any changes. (b) Stack reading $\mathbf{R}^t$ after an infinitesimal change of the length in the section $C_1$. The reading content has no change.

Since the magnitudes of $\Delta l(i)$ and $\Delta l(1)$ are the same as $\Delta A^\tau$, the ratio $\Delta l(i)_k/\Delta A^\tau$ (or $\Delta l(1)_k/\Delta A^\tau$) will be either one or zero depending on whether or not the symbol $C_i$ (or $C_1$) is the same as the $k_{th}$ symbol. This result can be expressed as

$$\frac{\partial R_k^t}{\partial A^\tau} = \delta_{ik} - \delta_{1k} \qquad \tau_i \le \tau < \tau_{i+1} \quad , \tag{A-10}$$

where $\delta_{ik}$ is *Kronecker* delta function.

Inserting Eq.(A-10) into Eq.(A-9) yields

$$\frac{\partial \boldsymbol{R}^t}{\partial \boldsymbol{W}} = \sum_{i=2}^{K} \left( (\delta_{ik} - \delta_{1k}) \sum_{\tau_i \le \tau < \tau_{i+1}} \frac{\partial A^\tau}{\partial \boldsymbol{W}} \right) \quad . \tag{A-11}$$

If we further assume that $K$=2, i.e. the current stack reading with length equal to 1 contains at most two sections of symbols; the approximation to Eq.(A-11) would be

$$\frac{\partial \boldsymbol{R}^t}{\partial \boldsymbol{W}} \approx (\delta_{ik} - \delta_{1k}) \sum_{\tau_2 \le \tau < t} \frac{\partial A^\tau}{\partial \boldsymbol{W}} \quad . \tag{A-12}$$

In this paper we also assume $\tau_2 = t$ and obtain

$$\frac{\partial \boldsymbol{R}^t}{\partial \boldsymbol{W}} \approx (\delta_{ik} - \delta_{1k}) \frac{\partial A^t}{\partial \boldsymbol{W}} \ . \tag{A-13}$$

This approximation implies that instead of considering the case where a section of the symbol in the stack is the cumulated results of many actions, the section $C_2$ is assumed to be generated by only one action.

The two approximations in Eqs.(A-12) and (A-13) are valid for the following two conditions: (1) when the action activity values are close to their saturation values 1 and -1 (or $|A^t|>0.5$). (2) or the total number of actions (i.e. the length of input string) is small. This corresponds to imposing a restriction on the learning strategy. During the initial stage of learning when the action activity values are far from their saturation values 1 and -1, short strings are used as the training examples and the string length is increased after the short strings have been learned.

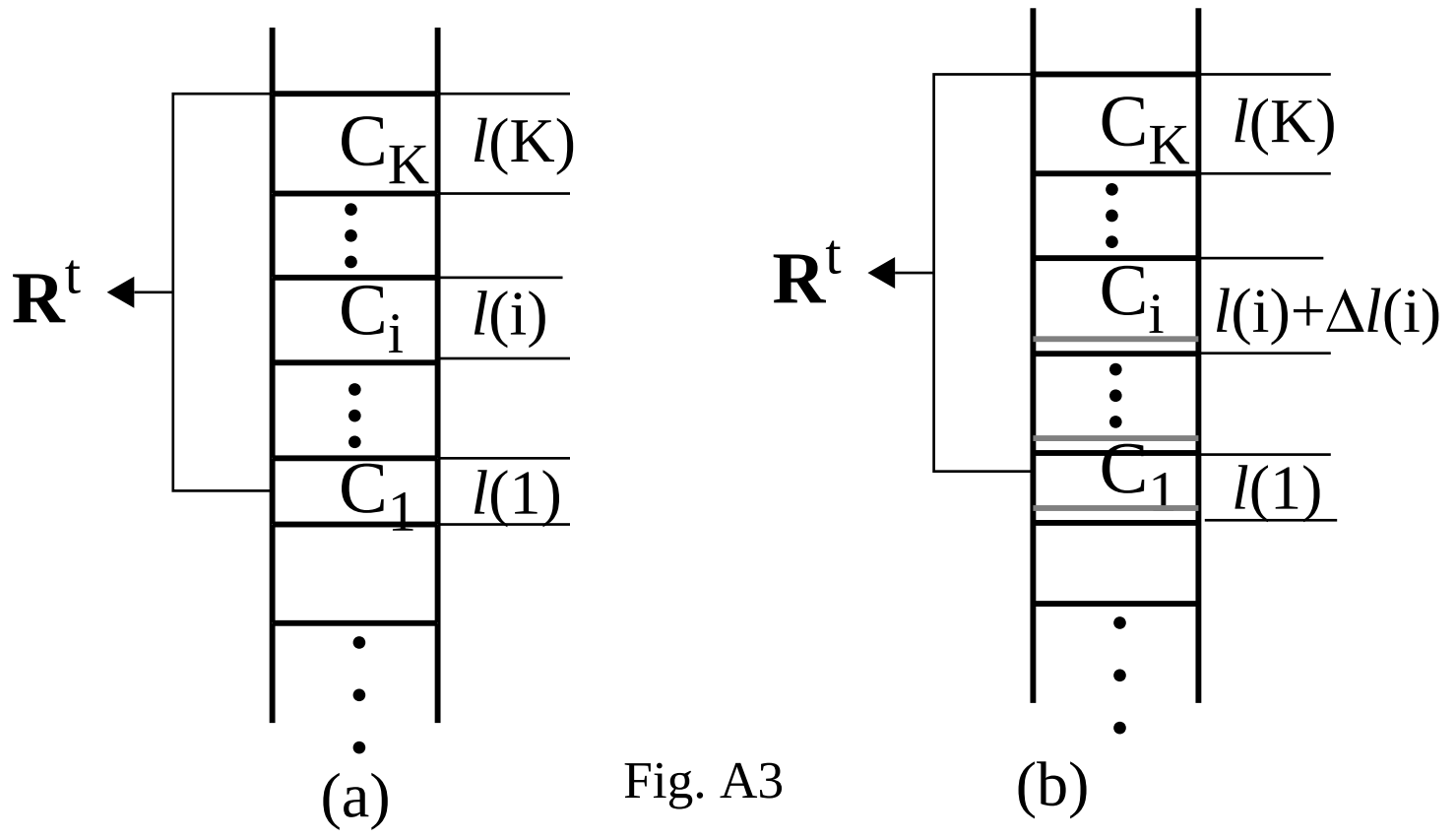

Fig. A3

Fig.A3 When the stack reading $\mathbf{R}^t$ has a fixed length = 1, the change of its content would include not only $\Delta$l(i), the change of the length of the symbol $C_i$, but also -$\Delta$l(1), the change in the length of symbol $C_1$.

# Appendix B

# Derivation of RTRL for the NNPDA

The forward-propagation recurrent learning algorithm known as Real Time Recurrent Learning (RTRL) can be derived by taking the derivative with respect to weights of the neural controller dynamics of Eqs. (23) and (24), and using Eq. (A-13) derived in Appendix A for the stack dynamics. For a complete appendix we first list these equations as follows

$$S_i^{t+1} = g\left(\sum_{j=1}^{N_S}\sum_{k=1}^{N_R}\sum_{l=1}^{N_I} W^s_{ijkl}\,(S_j^t R_k^t I_l^t) + \theta_i^s\right)$$

$$A^{t+1} = \sum_{J=1}^{2^{N_S}}\sum_{k=1}^{N_R}\sum_{l=1}^{N_I} W^a_{Jkl}\,(P_J^t R_k^t I_l^t) \quad , \qquad \text{(B-1)}$$

and

$$P_J^t = \prod_{m=1}^{N_S} (\delta_m S_m^t + (1-\delta_m)(1-S_m^t)) \quad . \qquad \text{(B-2)}$$

The derivative of the first equation in Eq. (B-1) is

$$\frac{\partial S_{i'}^{t+1}}{\partial W^s{}_{ijkl}} = g'_{i'}(t)\left(\delta_{ii'}(S_j^t R_k^t I_l^t) + \sum_{j'=1}^{N_s}\sum_{k'=1}^{N_I}\sum_{l'=1}^{N_I-1} W^s{}_{i'j'k'l'} I_{l'}^t\left(R_{k'}^t \frac{\partial S_{j'}^t}{\partial W^s{}_{ijkl}} + S_{j'}^t \frac{\partial R_{k'}^t}{\partial W^s{}_{ijkl}}\right)\right)$$

$$\frac{\partial S_{i'}^{t+1}}{\partial W^a{}_{Jkl}} = g'_{i'}(t)\sum_{j'=1}^{N_s}\sum_{k'=1}^{N_I}\sum_{l'=1}^{N_I-1} W^s{}_{i'j'k'l'} I_{l'}^t\left(R_{k'}^t \frac{\partial S_{j'}^t}{\partial W^a{}_{Jkl}} + S_{j'}^t \frac{\partial R_{k'}^t}{\partial W^a{}_{Jkl}}\right) \quad , \quad \text{(B-3)}$$

$$\frac{\partial S_{i'}^{t+1}}{\partial \theta^s{}_i} = g'_{i'}(t)\left(\delta_{ii'} + \sum_{j'=1}^{N_s}\sum_{k'=1}^{N_I}\sum_{l'=1}^{N_I-1} W^s{}_{i'j'k'l'} I_{l'}^t\left(R_{k'}^t \frac{\partial S_{j'}^t}{\partial \theta^s{}_i} + S_{j'}^t \frac{\partial R_{k'}^t}{\partial \theta^s{}_i}\right)\right)$$

where $g'_{i'}(t) = S_{i'}^{t+1}(1 - S_{i'}^{t+1})$ is the derivative of Sigmoid function. Similarly, the derivatives of second equation in Eq.(B-1) are written as

$$\frac{\partial A^{t+1}}{\partial W^s{}_{ijkl}} = \sum_{J'=1}^{2^{N_s}}\sum_{k'=1}^{N_I}\sum_{l'=1}^{N_I-1} W^a{}_{J'k'l'} I_{l'}^t\left(R_{k'}^t \frac{\partial P_{J'}^t}{\partial W^s{}_{ijkl}} + P_{J'}^t \frac{\partial R_{k'}^t}{\partial W^s{}_{ijkl}}\right)$$

$$\frac{\partial A^{t+1}}{\partial W^a{}_{Jkl}} = P_J^t R_k^t I_l^t + \sum_{J'=1}^{2^{N_s}}\sum_{k'=1}^{N_I}\sum_{l'=1}^{N_I-1} W^a{}_{J'k'l'} I_{l'}^t\left(R_{k'}^t \frac{\partial P_{J'}^t}{\partial W^a{}_{Jkl}} + P_{J'}^t \frac{\partial R_{k'}^t}{\partial W^a{}_{Jkl}}\right). \quad \text{(B-4)}$$

$$\frac{\partial A^{t+1}}{\partial \theta^s{}_i} = \sum_{J'=1}^{2^{N_s}}\sum_{k'=1}^{N_I}\sum_{l'=1}^{N_I-1} W^a{}_{J'k'l'} I_{l'}^t\left(R_{k'}^t \frac{\partial P_{J'}^t}{\partial \theta^s{}_i} + P_{J'}^t \frac{\partial R_{k'}^t}{\partial \theta^s{}_i}\right)$$

To complete the derivation we need two more relations which are obtained from the derivative of Eq.(B-2) and the derivative of stack reading in Eq.(20)

$$\frac{\partial P_J^t}{\partial W} = P_J^t \sum_{m=1}^{N_s} \frac{(\delta_m - S_m^t)}{S_m^t(1 - S_m^t)} \frac{\partial S_m^t}{\partial W} \quad , \quad \text{(B-5)}$$

$$\frac{\partial R_{k'}^t}{\partial W} \approx (\delta_{k'r_1^t} - \delta_{k'r_2^t}) \frac{\partial A^t}{\partial W}$$

where $r_1^t$ and $r_2^t$ are the ordinal numbers of neurons that represent the top and the bottom symbols respectively in the reading $\boldsymbol{R}^t$. The initial conditions for all the derivatives in Eqs.(B-3) to (B-5) are set to zero.